\documentclass[]{fairmeta}

\usepackage{algorithm,algorithmic}
\usepackage{amsmath}
\usepackage{amssymb}
\usepackage{txfonts}
\usepackage{graphicx}
\usepackage{tabularx}
\usepackage{booktabs} 
\usepackage{subcaption}
\usepackage{url}
\usepackage{CJKutf8} 

\newcommand{\nlang}{8 }
\newcommand{\ndata}{4,501 }

\usepackage{booktabs}
\usepackage{multirow}

\usepackage{makecell}

\usepackage{xspace}
\newcommand{\dataset}{\textsc{Multi-IF}\xspace}
\usepackage{xcolor}
\usepackage{tcolorbox}

\allowdisplaybreaks[4]

\title{Multi-IF: Benchmarking LLMs on Multi-Turn and Multilingual Instructions Following}

\author{Yun He*}
\author{Di Jin*}
\author{Chaoqi Wang*}
\author{Chloe Bi*}
\author{Karishma Mandyam}
\author{Hejia Zhang}
\author{Chen Zhu}
\author{Ning Li}
\author{Tengyu Xu}
\author{Hongjiang Lv}
\author{Shruti Bhosale}
\author{Chenguang Zhu}
\author{Karthik Abinav Sankararaman}
\author{Eryk Helenowski}
\author{Melanie Kambadur}
\author{Aditya Tayade}
\author{Hao Ma}
\author{Han Fang}
\author{Sinong Wang}
\affiliation{Meta GenAI}

\contribution[*]{Co-first Author}

\abstract{
Large Language Models (LLMs) have demonstrated impressive capabilities in various tasks, including instruction following, which is crucial for aligning model outputs with user expectations. However, evaluating LLMs' ability to follow instructions remains challenging due to the complexity and subjectivity of human language. Current benchmarks primarily focus on single-turn, monolingual instructions, which do not adequately reflect the complexities of real-world applications that require handling multi-turn and multilingual interactions. To address this gap, we introduce \dataset, a new benchmark designed to assess LLMs' proficiency in following multi-turn and multilingual instructions. \dataset, which utilizes a hybrid framework combining LLM and human annotators, expands upon the IFEval by incorporating multi-turn sequences and translating the English prompts into another 7 languages, resulting in a dataset of \ndata multilingual conversations, where each has three turns. Our evaluation of 14 state-of-the-art LLMs on \dataset reveals that it presents a significantly more challenging task than existing benchmarks. All the models tested showed a higher rate of failure in executing instructions correctly with each additional turn. For example, o1-preview drops from 0.877 at the first turn to 0.707 at the third turn in terms of average accuracy over all languages. Moreover, languages with non-Latin
scripts (Hindi, Russian, and Chinese) generally exhibit higher error rates, suggesting potential limitations in the models’ multilingual capabilities. We release \dataset prompts\footnote{\url{https://huggingface.co/datasets/facebook/Multi-IF}} and the evaluation code\footnote{\url{https://github.com/facebookresearch/Multi-IF}} base to encourage further research in this critical area.
}

\date{\today}
\correspondence{Yun He at \email{yunhe2019@meta.com}}

\begin{document}

\maketitle

\section{Introduction}\label{section:intro}

\begin{figure}[h]
  \centering
  \setlength{\abovecaptionskip}{0.0cm}
  \setlength{\belowcaptionskip}{0.0cm}
  \includegraphics[width=0.9\linewidth]{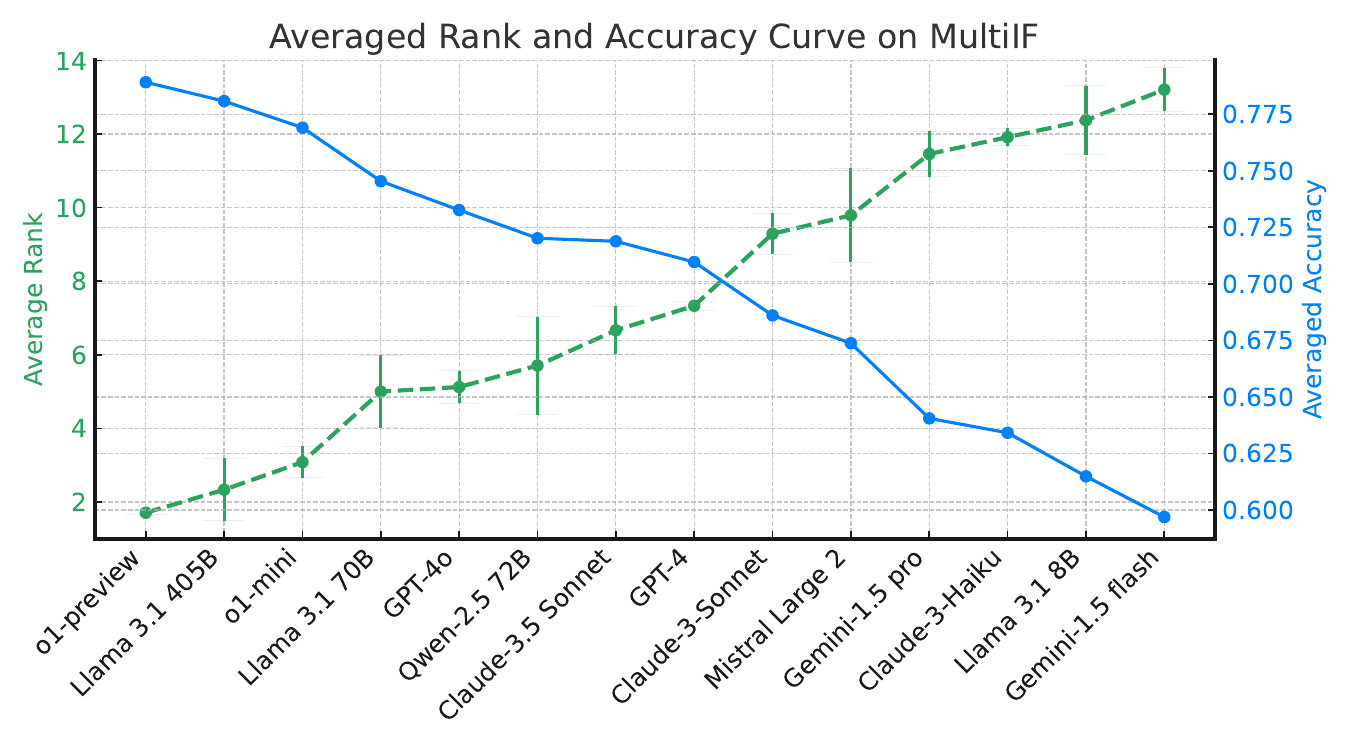}
  \caption{Overall Performance of Large Language Models on \dataset. We conduct our comparison mainly by evaluating the average scores of the models across all languages and across all turns.}
\label{fig:overall_results}
\end{figure}


Large language models (LLMs) have proven their remarkable abilities in addressing various kinds of tasks in both zero-shot and few-shots settings~\citep{achiam2023gpt,reid2024gemini,dubey2024llama}. Among all capabilities, instruction following is usually considered as one of the most crucial tasks for LLM applications, since it determines how well the model outputs align with user requirements. In real-world use of LLMs, user inputs usually come with various kinds of instructions to specify constraints on the model output. Discrepancies or misunderstandings in following instructions can lead to unintended outputs, dissatisfying users or even causing severe catastrophes in crucial scenarios like healthcare or autonomous systems. Therefore, ensuring that LLMs can consistently adhere to given directives is among the top priorities. To develop the instruction following capability, we first need to build a reliable and comprehensive framework to evaluate a model's ability to follow instructions. 

However, evaluating the instruction following ability of LLMs is a non-trivial task~\citep{sun2023evaluating}. The biggest challenges lie in the complexity, subjectivity, and ambiguity of human language. The same text can be interpreted differently, leading to varying judgments when evaluating whether a model has followed instructions, especially when instructions are subjective, such as ``write in a humorous tone''. To circumvent this challenge, a line of work, such as IFEval~\citep{zhou2023instruction}, focused on evaluating a LLM's ability to follow ``verifiable instructions'', which are defined as instructions amenable to objective verification of compliance. Examples of such instructions can be ``the response should be in three paragraphs'', ``the response should be in more than 300 words'', etc. Based on these verifiable instructions, we can build an automatic, quantifiable, and accurate evaluation framework to assess a LLM's ability to follow directions. 

Moreover, in real-world user-AI conversation settings, a majority of interactions happen in a long multi-turn format, e.g., 10+ turns. Unfortunately, almost all existing instruction following benchmarks consist of only single-turn instructions~\citep{zhou2023instruction,jiang2023followbench,qin2024infobench,wen2024benchmarking}, causing insufficient evaluation of the LLMs’ ability to follow multi-turn instructions. Besides, a majority of these benchmarks are in single languages, such as English only~\citep{li2023alpacaeval,zheng2024judging}, Chinese only~\citep{liu2023alignbench}, etc. These drawbacks make them insufficient to evaluate a real-world LLM based chatbot or service that serves billions of users across the world, such as Meta AI, Gemini, ChatGPT, etc. 

\begin{figure}[h]
  \centering
  \setlength{\abovecaptionskip}{0.2cm}
  \setlength{\belowcaptionskip}{0.0cm}
  \includegraphics[width=0.95\linewidth]{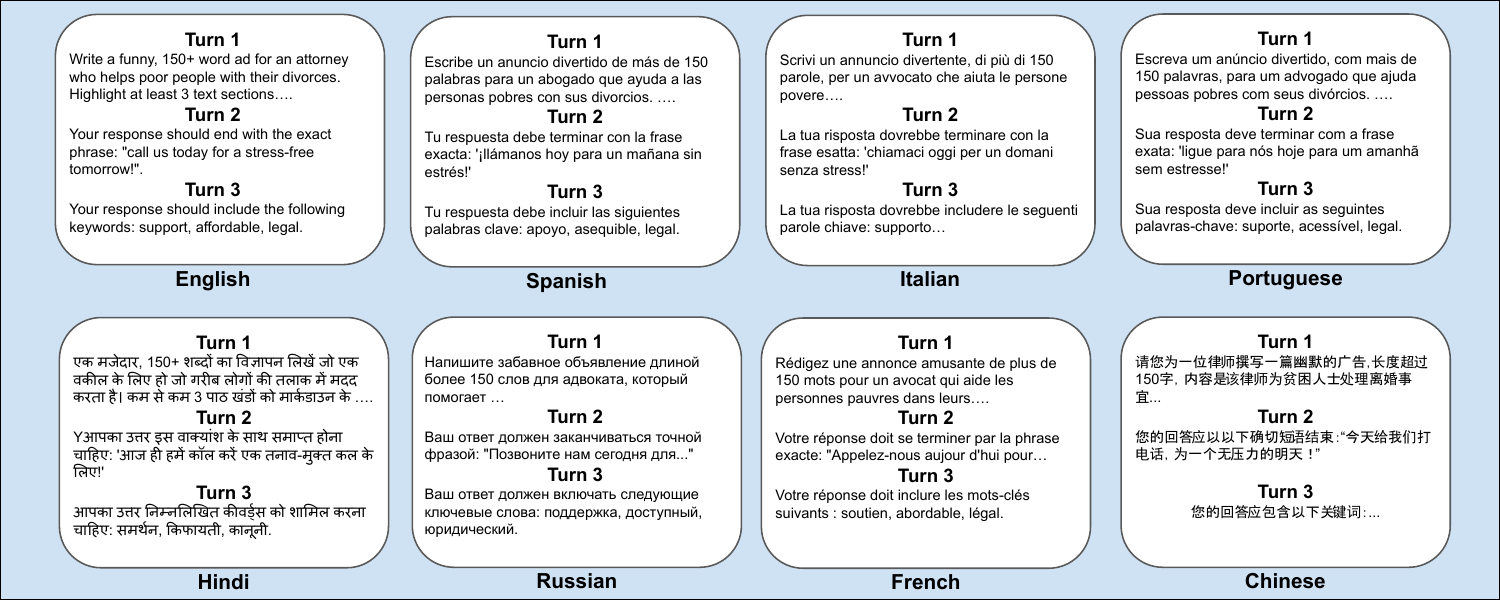}
  \caption{Examples of Multi-Instruction Following (\dataset) tasks in various languages. Each block contains a three-turn interaction for writing a creative advertisement for a divorce attorney (\textbf{instruction is synthetically generated, not real user data}), with specific instructions provided in English, Spanish, Italian, Portuguese, Hindi, Russian, French, and Chinese. The tasks involve word count limits, inclusion of exact phrases, and use of specific keywords related to the attorney's services.}
\label{fig: multi_if_examples}
\end{figure}

In this work, we introduce \dataset, a new multi-turn and multilingual benchmark for evaluating the proficiency of LLMs in instruction following. Based on IFEval~\citep{zhou2023instruction}, a dataset of English single-turn prompts with verifiable instructions, we propose a framework to systematically expand it into a dataset of multi-turn and multilingual prompts. Specifically, we first expand each single turn user prompt into a multi-turn user prompt via random sampling and LLM based prompt revision and expansion. After that, we remove conflicted instructions between turns by prompting an LLM for automatic filtering followed by human auditing. Finally, the English-only dataset is translated into other 7 languages: French, Russian, Hindi, Italian, Portuguese, Spanish and Chinese. The translation process is first performed by a LLM and then audited by professional human annotators. Overall, we created \ndata multi-turn prompts in \nlang languages. Figure \ref{fig: multi_if_examples} shows one example for each language. 

Based on \dataset, we report the evaluation results of several state-of-the-art (SOTA) LLMs, including Llama 3.1~\citep{dubey2024llama}, GPT-4o~\citep{achiam2023gpt}, and OpenAI o1~\citep{openaio1}. Our findings indicate that:

\begin{itemize}
    \item \dataset presents a significantly more challenging instruction-following benchmark than existing ones, such as IFEval, highlighting considerable room for improvement in this area.
    \item LLMs exhibit difficulty maintaining high accuracy in following instructions as the number of turns increases. Our results demonstrate a growing failure rate among LLMs to follow instructions correctly with each additional turn. Moreover, strong reasoning capabilities are beneficial for recovering from errors made in prior turns.
    \item We observed that as the number of turns increases, LLMs increasingly forget to adhere to instructions that were successfully executed in previous turns, which contributes to the performance degradation.
    \item A noteworthy observation is the reduced performance of all evaluated models in non-English languages compared to English, indicating a significant opportunity for enhancement in multilingual instruction-following capabilities.
\end{itemize}

\section{Related Work}


\textbf{Evaluation Methodologies.} Evaluating the instruction-following capabilities of LLMs is a critical area of research, with approaches broadly classified into three categories: human evaluation, model-based evaluation, and rule/script-based evaluation. \textit{Human evaluation} involves human annotators assessing the performance of LLMs on various tasks. Studies such as \citet{ouyang2022training} and \citet{taori2023stanford} employ human evaluators to rate the quality of model responses to complex instructions. While this approach allows for nuanced judgment over a wide range of tasks, it is time-consuming, expensive, and can suffer from inconsistencies among annotators. Reproducibility is also a challenge due to variations in annotator interpretations and criteria. Other works like \citet{zheng2023judging} and \citet{kocmi2023large} have explored scaling human evaluation, but the issues of cost and inconsistency remain.  With advancements in LLMs, \textit{model-based rvaluation} has become a popular alternative. This approach leverages a strong LLM (e.g., GPT-4) to assess the outputs of other models~\citep{jin2023data,xu2023wizardlm,liu2023g,jiang2023followbench,qin2024infobench,chen2024mj,chang2024survey,wang2023beyond,wen2024benchmarking,xia2024fofo}. While this method is scalable and less resource-intensive than human evaluation, it heavily depends on the correctness and objectivity of the judge model. Biases and errors inherent in the judge model can introduce inaccuracies in the evaluation~\citep{wang2023large}. Additionally, the black-box nature of some judge models can limit transparency. \textit{Rule/Script-based evaluation}  employs predefined rules or scripts to objectively verify the outputs of LLMs against verifiable instructions~\citep{sun2023evaluating,zhou2023instruction,he2024can}. This method provides standardized and scalable evaluation without human or model biases. However, current benchmarks in this category are often limited to single-turn interactions and predominantly in English, restricting their applicability in real-world multilingual and multi-turn conversational settings.

\textbf{Multi-lingual and Multi-turn Benchmarks.}  Addressing the need for broader evaluation, several \textit{multilingual and multi-turn benchmarks} have been developed for multilingual and conversational tasks. The XTREME benchmark~\citep{siddhant2020xtreme} and its successor XTREME-R~\citep{ruder2021xtreme} evaluate cross-lingual generalization across diverse languages. For dialogue systems, datasets like MultiWOZ~\citep{budzianowski2018multiwoz} and DailyDialog~\citep{li2017dailydialog} focus on multi-turn conversations but are limited to English and specific domains. Our work builds upon the rule/script-based evaluation method and extends it to a multi-turn and multilingual benchmark. By incorporating multiple languages and conversational turns, we provide a more comprehensive evaluation of LLMs' instruction-following capabilities, reflecting the needs of billions of users worldwide in realistic conversational scenarios. \textit{Evaluation of Multi-Turn Dialogue Systems} presents unique challenges~\citep{mehri2020usr}. Traditional metrics like BLEU~\citep{papineni2002bleu} and ROUGE~\citep{lin2004rouge} are insufficient for capturing the quality of conversations~\citep{liu2016not}. Recent works have proposed learned metrics such as BLEURT~\citep{sellam2020bleurt} and BERTScore~\citep{zhang2020bertscore} for more accurate assessments. User simulators have also been explored to evaluate dialogue systems in a scalable manner~\citep{asri2016sequence,tseng2021transferable}. Our benchmark leverages these insights to create an evaluation framework suited for multi-turn interactions.

\textbf{Fairness and Bias in Multilingual Models.} The expansion into multilingual evaluation raises concerns about fairness and bias in LLMs across different languages~\citep{blasi2021systematic,wu2020are}. Ensuring equitable performance in instruction following requires careful consideration of linguistic and cultural nuances. Our benchmark aims to highlight these issues by including a diverse set of languages, thus encouraging the development of models that perform consistently across linguistic boundaries.

\section{\dataset}
\begin{figure}[h]
  \centering
  \setlength{\abovecaptionskip}{0.0cm}
  \setlength{\belowcaptionskip}{0.0cm}
  \includegraphics[width=0.80\linewidth]{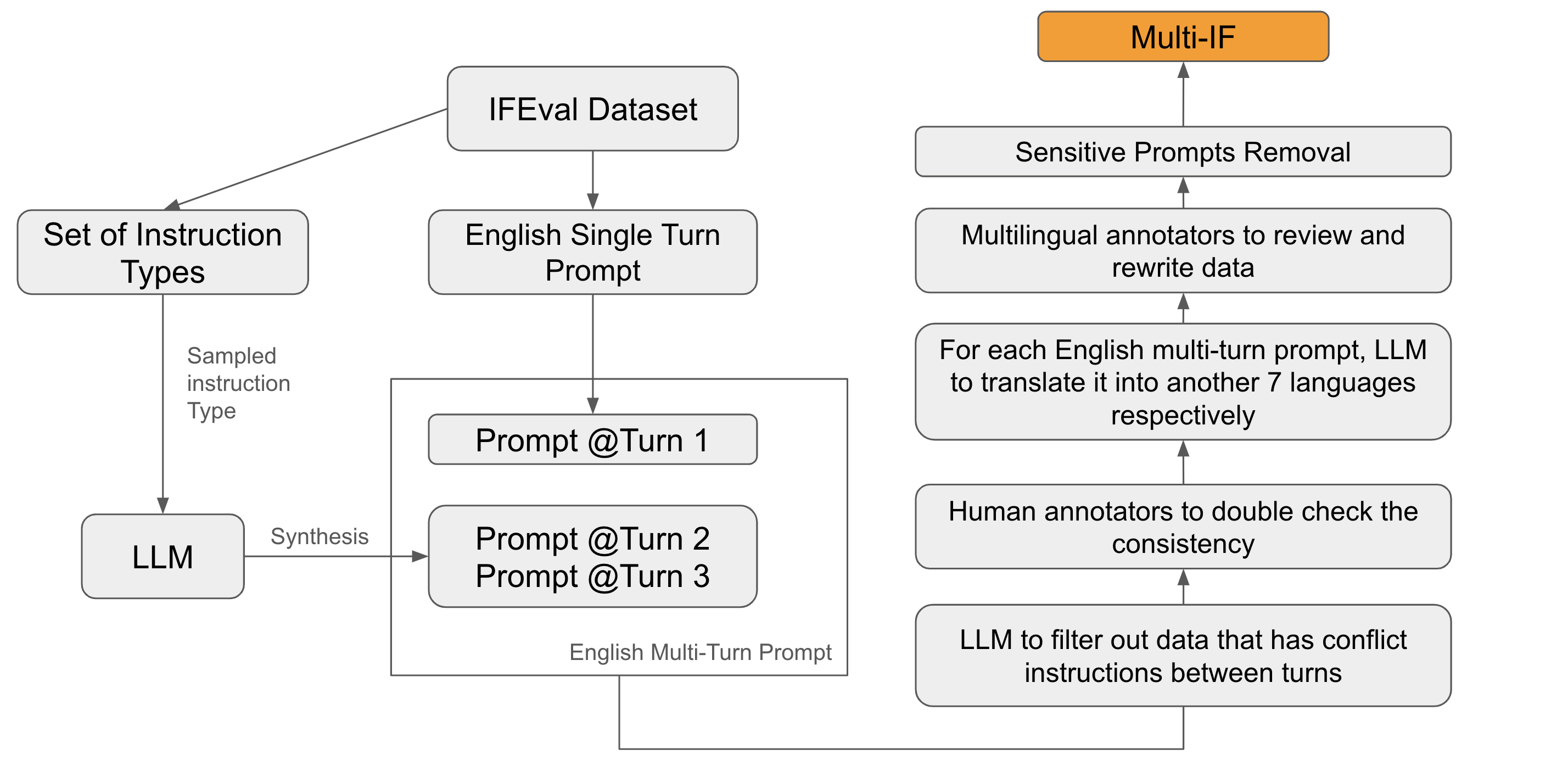}
  \caption{Overall framework to build \dataset.}
\label{fig: multiif_method}
\end{figure}

As shown in Figure~\ref{fig: multiif_method}, we propose a framework to expand a dataset of English single-turn prompts to a dataset of multilingual and multi-turn prompts. We applied it to expand IFEval ~\citep{zhou2023instruction} into a new and much more challenging benchmark called \dataset. 

\subsection{Multi-turn Expansion}
\begin{figure}[t]
  \centering
  \setlength{\abovecaptionskip}{0.0cm}
  \setlength{\belowcaptionskip}{0.0cm}
  \includegraphics[width=0.80\linewidth]{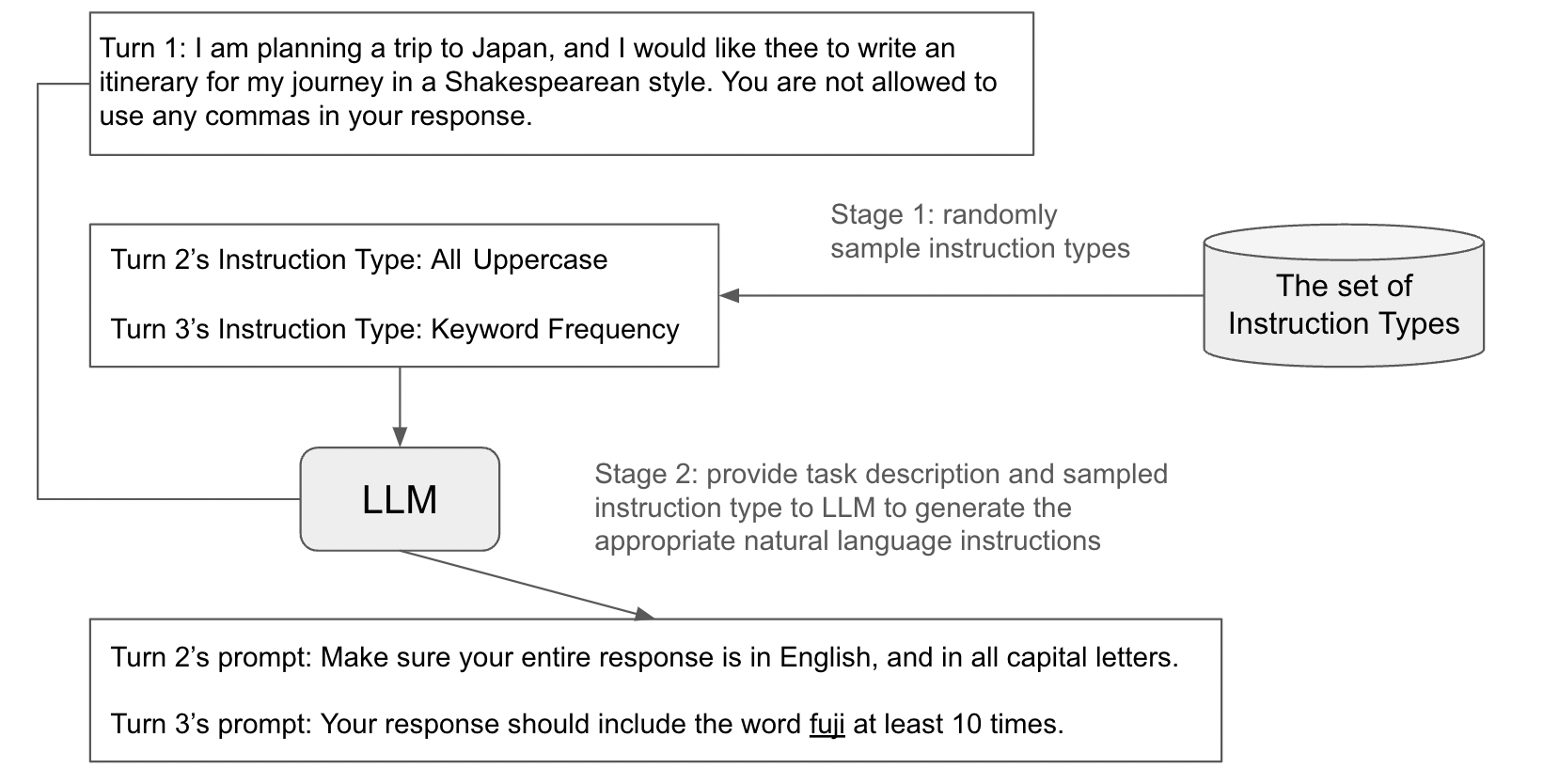}
  \caption{Two stage method to expand a single-turn prompt into multi-turn prompts.}
\label{fig: multiif_method_detail}
\end{figure}

This subsection will introduce how we generate more following user prompts given a single-turn prompt in two stages. By default, we generate 2 more following user prompts out of the first one to form a three-turn conversation. For English, we use the original prompt in IFEval as the first turn. As illustrated in the Figure \ref{fig: multiif_method_detail}, we first randomly sample (random sample to diversify the instruction types as more as possible) one out of the 30 pre-defined instruction types in IFEval dataset for each following turn. For example, for turn 2 in Figure \ref{fig: multiif_method_detail}, we randomly sampled the instruction type ``All Uppercase''. Then, we prompt an LLM (Llama 3.1 405B) to generate an appropriate natural language user prompt given the sampled instruction type and all the previous turns' prompts. For example, for turn 2, we use the LLM to turn the instruction type ``All Uppercase'' into the natural language instruction of ``Make sure your entire response is in English, and in all capital letters''.

\subsection{Conflict Instructions Removal}
To ensure the integrity and consistency of instructions in multi-turn prompts, we have implemented a two-step conflict resolution process. In the first step, we utilize Llama 3.1 405B to scan through the multi-turn prompts and identify any conflicting instructions across turns. For example, a conflict might occur if the first turn requests a short summary while the second turn requires a response of at least 800 words. In the second step, human verification is conducted to filter out any prompts where instructions contradict with each other across different turns. By employing this two-stage process, we guarantee that the remaining data contain multi-turn prompts with compatible and follow-able instructions, enabling LLMs to consistently adhere to the directives from the first to the last turn.

\subsection{Multilingual Translation}
Given the golden set of English multi-turn prompts, we use Llama 3.1 405B to translate them into other 7 languages: French, Hindi, Italian, Portuguese, Chinese, Spanish and Russian. Note that some instructions can only apply to western Languages, e.g., English, such as ``the response should be in all capital letters", it is not reasonable to translate such instructions to other non-Western languages like Chinese. According to this limit, we refrain from translating certain instructions for a subset of languages. 

After that, multilingual annotators evaluate all translations to ensure that (1) the translation remains faithful to the original English prompt, and (2) the translation aligns with the style of a native speaker. Additionally, these annotators undertake the task of revising the translations to meet these two criteria. There are 15\% translations get rewritten in average of all languages. 

\subsection{Sensitive Prompts Removal}
The localized translation sometimes will generate some sensitive content in the prompts. To ensure the removal of sensitive content related to politics, religious sensitivity, and human relationships from our dataset, we initially employ a LLM to scan and flag potentially sensitive prompts. Subsequently, human annotators review these flagged prompts to accurately identify and eliminate those containing actual sensitive content.

\subsection{Examples and Statistics}

\begin{figure}[htp]
    \centering
    \includegraphics[width=\textwidth]{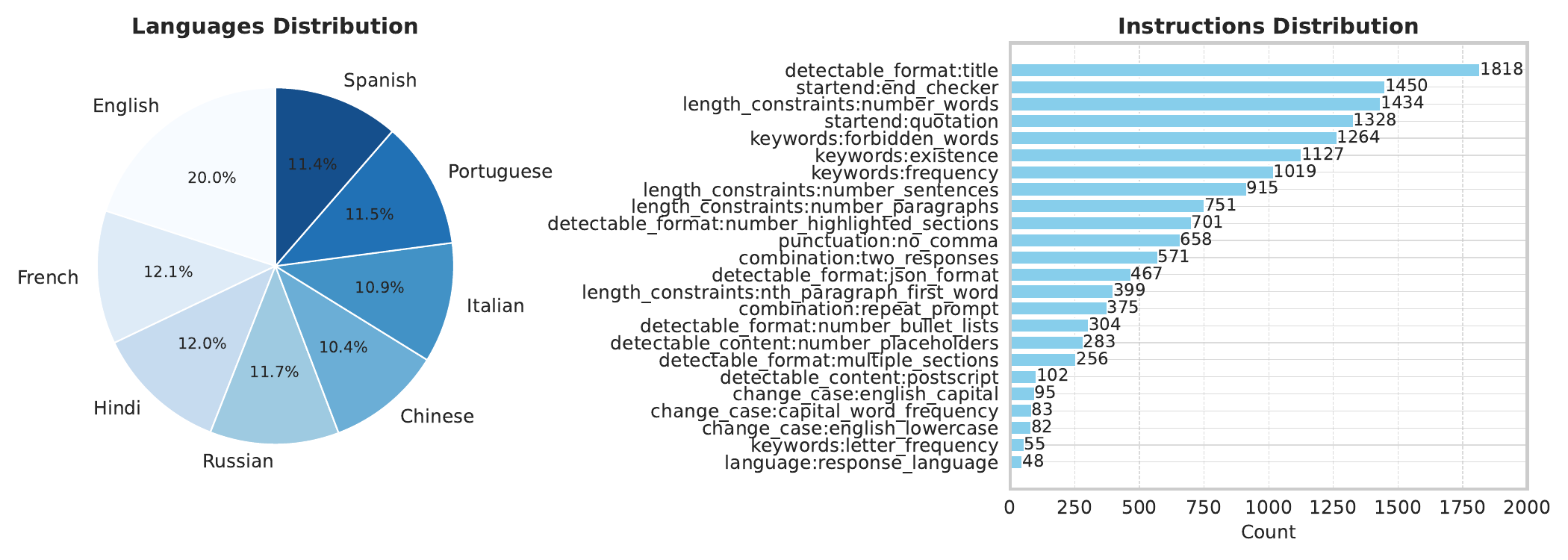}
    \caption{The distribution of languages (left) and instruction types (right) in \dataset. }
    \label{fig: distribution}
\end{figure}

There are 4,501 conversations in the benchmark, each of which contains three turns. We show the distribution of languages and instruction category in Figure \ref{fig: distribution}. English set has more prompts than the other languages due to two reasons: (1) We expand each single-turn English prompt into two multi-turn prompts, but we only translate a part of them into other languages because some instructions can only apply to western Languages; (2) some multilingual multi-turn prompts are dropped during the sensitive prompts removal;


\begin{table}[htbp]
  \small
  \centering
    \caption{Average accuracy scores of several leading LLMs on various languages. The average of the four accuracy scores is reported: instruction-level strict accuracy, conversation-level strict accuracy, instruction-level loose accuracy, and conversation-level loose accuracy. ``Average'' refers to the averaged accuracy across all languages.}\label{tab:evaluation-results-accuracy-main}
    \begin{tabularx}{\textwidth}{l*{10}{c}}
    \toprule
    \textbf{Turn 1} & \multicolumn{1}{c}{\textbf{Average}} & \multicolumn{1}{c}{\textbf{English}} & \multicolumn{1}{c}{\textbf{French}} & \multicolumn{1}{c}{\textbf{Russian}} & \multicolumn{1}{c}{\textbf{Hindi}} & \multicolumn{1}{c}{\textbf{Italian}} & \multicolumn{1}{c}{\textbf{Portuguese}} & \multicolumn{1}{c}{\textbf{Spanish}} & \multicolumn{1}{c}{\textbf{Chinese}} \\ \hline
o1-preview & 0.877 & 0.856 & 0.898 & 0.835 & 0.871 & 0.891 & 0.895 & 0.912 & 0.858 & \\ 
o1-mini & 0.853 & 0.836 & 0.882 & 0.815 & 0.842 & 0.873 & 0.868 & 0.886 & 0.824 & \\ 
GPT-4o & 0.843 & 0.874 & 0.853 & 0.789 & 0.812 & 0.878 & 0.858 & 0.876 & 0.805 & \\ 
GPT-4 & 0.815 & 0.860 & 0.842 & 0.789 & 0.718 & 0.840 & 0.832 & 0.850 & 0.786 & \\ 
Llama 3.1 405B & 0.854 & 0.907 & 0.868 & 0.801 & 0.825 & 0.864 & 0.876 & 0.871 & 0.817 & \\ 
Llama 3.1 70B & 0.826 & 0.890 & 0.837 & 0.783 & 0.759 & 0.862 & 0.847 & 0.847 & 0.783 & \\ 
Llama 3.1 8B & 0.688 & 0.801 & 0.693 & 0.617 & 0.587 & 0.702 & 0.720 & 0.743 & 0.640 & \\ 
Gemini-1.5 pro & 0.758 & 0.835 & 0.763 & 0.743 & 0.719 & 0.745 & 0.755 & 0.756 & 0.750 & \\ 
Gemini-1.5 flash & 0.725 & 0.775 & 0.715 & 0.686 & 0.691 & 0.766 & 0.727 & 0.722 & 0.717 & \\ 
Claude-3.5 Sonnet & 0.817 & 0.876 & 0.828 & 0.777 & 0.782 & 0.805 & 0.837 & 0.859 & 0.773 & \\ 
Claude-3-Sonnet & 0.782 & 0.831 & 0.774 & 0.763 & 0.754 & 0.803 & 0.801 & 0.818 & 0.714 & \\ 
Claude-3-Haiku & 0.729 & 0.779 & 0.727 & 0.719 & 0.674 & 0.752 & 0.754 & 0.766 & 0.660 & \\ 
Qwen-2.5 72B & 0.837 & 0.881 & 0.869 & 0.822 & 0.681 & 0.865 & 0.867 & 0.882 & 0.831 & \\ 
Mistral Large 2 & 0.805 & 0.835 & 0.837 & 0.772 & 0.678 & 0.817 & 0.840 & 0.851 & 0.808 & \\ 

    \midrule
    \textbf{Turn 2} & \multicolumn{1}{l}{\textbf{Average}} & \multicolumn{1}{l}{\textbf{English}} & \multicolumn{1}{l}{\textbf{French}} & \multicolumn{1}{l}{\textbf{Russian}} & \multicolumn{1}{l}{\textbf{Hindi}} & \multicolumn{1}{l}{\textbf{Italian}} & \multicolumn{1}{l}{\textbf{Portuguese}} & \multicolumn{1}{l}{\textbf{Spanish}} & \multicolumn{1}{l}{\textbf{Chinese}} \\ \hline
o1-preview & 0.783 & 0.834 & 0.820 & 0.633 & 0.775 & 0.819 & 0.811 & 0.804 & 0.766 & \\ 
o1-mini & 0.772 & 0.802 & 0.800 & 0.684 & 0.763 & 0.812 & 0.782 & 0.809 & 0.728 & \\ 
GPT-4o & 0.724 & 0.784 & 0.745 & 0.601 & 0.708 & 0.760 & 0.740 & 0.749 & 0.705 & \\ 
GPT-4 & 0.705 & 0.756 & 0.752 & 0.619 & 0.634 & 0.731 & 0.734 & 0.729 & 0.685 & \\ 
Llama 3.1 405B & 0.782 & 0.843 & 0.822 & 0.692 & 0.754 & 0.793 & 0.801 & 0.800 & 0.748 & \\ 
Llama 3.1 70B & 0.742 & 0.814 & 0.774 & 0.629 & 0.707 & 0.781 & 0.778 & 0.757 & 0.697 & \\ 
Llama 3.1 8B & 0.615 & 0.720 & 0.663 & 0.472 & 0.540 & 0.631 & 0.645 & 0.678 & 0.570 & \\ 
Gemini-1.5 pro & 0.624 & 0.710 & 0.646 & 0.565 & 0.578 & 0.625 & 0.610 & 0.641 & 0.613 & \\ 
Gemini-1.5 flash & 0.570 & 0.644 & 0.602 & 0.420 & 0.534 & 0.628 & 0.571 & 0.574 & 0.591 & \\ 
Claude-3.5 Sonnet & 0.705 & 0.780 & 0.724 & 0.632 & 0.679 & 0.705 & 0.707 & 0.736 & 0.679 & \\ 
Claude-3-Sonnet & 0.680 & 0.753 & 0.716 & 0.616 & 0.672 & 0.691 & 0.685 & 0.702 & 0.609 & \\ 
Claude-3-Haiku & 0.632 & 0.693 & 0.654 & 0.594 & 0.576 & 0.644 & 0.654 & 0.669 & 0.571 & \\ 
Qwen-2.5 72B & 0.715 & 0.764 & 0.769 & 0.664 & 0.582 & 0.753 & 0.738 & 0.747 & 0.700 & \\ 
Mistral Large 2 & 0.669 & 0.734 & 0.721 & 0.598 & 0.541 & 0.698 & 0.678 & 0.705 & 0.675 & \\

    \midrule
    \textbf{Turn 3} & \multicolumn{1}{l}{\textbf{Average}} & \multicolumn{1}{l}{\textbf{English}} & \multicolumn{1}{l}{\textbf{French}} & \multicolumn{1}{l}{\textbf{Russian}} & \multicolumn{1}{l}{\textbf{Hindi}} & \multicolumn{1}{l}{\textbf{Italian}} & \multicolumn{1}{l}{\textbf{Portuguese}} & \multicolumn{1}{l}{\textbf{Spanish}} & \multicolumn{1}{l}{\textbf{Chinese}} \\ \hline
o1-preview & 0.707 & 0.773 & 0.738 & 0.531 & 0.709 & 0.759 & 0.714 & 0.733 & 0.703 & \\ 
o1-mini & 0.681 & 0.742 & 0.716 & 0.576 & 0.668 & 0.701 & 0.708 & 0.714 & 0.627 & \\ 
GPT-4o & 0.631 & 0.701 & 0.653 & 0.501 & 0.621 & 0.647 & 0.645 & 0.645 & 0.631 & \\ 
GPT-4 & 0.609 & 0.678 & 0.653 & 0.490 & 0.537 & 0.633 & 0.639 & 0.634 & 0.608 & \\ 
Llama 3.1 405B & 0.707 & 0.786 & 0.753 & 0.587 & 0.677 & 0.740 & 0.727 & 0.716 & 0.670 & \\ 
Llama 3.1 70B & 0.668 & 0.749 & 0.718 & 0.519 & 0.640 & 0.710 & 0.696 & 0.689 & 0.622 & \\ 
Llama 3.1 8B & 0.542 & 0.641 & 0.590 & 0.384 & 0.466 & 0.570 & 0.586 & 0.604 & 0.495 & \\ 
Gemini-1.5 pro & 0.540 & 0.641 & 0.554 & 0.467 & 0.478 & 0.544 & 0.542 & 0.550 & 0.540 & \\ 
Gemini-1.5 flash & 0.496 & 0.563 & 0.512 & 0.346 & 0.455 & 0.539 & 0.521 & 0.507 & 0.522 & \\ 
Claude-3.5 Sonnet & 0.634 & 0.725 & 0.649 & 0.529 & 0.614 & 0.639 & 0.632 & 0.653 & 0.630 & \\ 
Claude-3-Sonnet & 0.596 & 0.678 & 0.644 & 0.491 & 0.590 & 0.615 & 0.598 & 0.616 & 0.532 & \\ 
Claude-3-Haiku & 0.542 & 0.612 & 0.554 & 0.485 & 0.484 & 0.549 & 0.579 & 0.567 & 0.504 & \\ 
Qwen-2.5 72B & 0.609 & 0.672 & 0.645 & 0.527 & 0.497 & 0.648 & 0.645 & 0.626 & 0.608 & \\ 
Mistral Large 2 & 0.548 & 0.638 & 0.594 & 0.458 & 0.439 & 0.563 & 0.560 & 0.576 & 0.552 & \\

    \bottomrule
    \end{tabularx}
\end{table}

    \begin{table}[htbp]
      \small
      \centering
      \caption{Ranks of several leading LLMs on various languages, which are computed based on the accuracy in Table~\ref{tab:evaluation-results-accuracy-main}. ``Average'' refers to the average ranks across all languages.} \label{tab:evaluation-results-ranks-main}
      \begin{tabularx}{\textwidth}{l*{10}{c}}
    
 \toprule 
 \textbf{Turn 1} & \textbf{Avg.} & \textbf{English} & \textbf{French} & \textbf{Russian} & \textbf{Hindi} & \textbf{Italian} & \textbf{Portuguese} & \textbf{Spanish} & \textbf{Chinese} \\ \midrule 
o1-preview & 1.75 & 7 & 1 & 1 & 1 & 1 & 1 & 1 & 1 & \\ 
o1-mini & 3.25 & 8 & 2 & 3 & 2 & 3 & 3 & 2 & 3 & \\ 
GPT-4o & 4.50 & 5 & 5 & 5 & 4 & 2 & 5 & 4 & 6 & \\ 
GPT-4 & 7.25 & 6 & 6 & 6 & 9 & 7 & 9 & 8 & 7 & \\ 
Llama 3.1 405B & 3.50 & 1 & 4 & 4 & 3 & 5 & 2 & 5 & 4 & \\ 
Llama 3.1 70B & 6.38 & 2 & 7 & 7 & 6 & 6 & 6 & 9 & 8 & \\ 
Llama 3.1 8B & 13.62 & 12 & 14 & 14 & 14 & 14 & 14 & 13 & 14 & \\ 
Gemini-1.5 pro & 10.62 & 9 & 11 & 11 & 8 & 13 & 11 & 12 & 10 & \\ 
Gemini-1.5 flash & 12.38 & 14 & 13 & 13 & 10 & 11 & 13 & 14 & 11 & \\ 
Claude-3.5 Sonnet & 7.25 & 4 & 9 & 8 & 5 & 9 & 8 & 6 & 9 & \\ 
Claude-3-Sonnet & 10.00 & 11 & 10 & 10 & 7 & 10 & 10 & 10 & 12 & \\ 
Claude-3-Haiku & 12.25 & 13 & 12 & 12 & 13 & 12 & 12 & 11 & 13 & \\ 
Qwen-2.5 72B & 4.00 & 3 & 3 & 2 & 11 & 4 & 4 & 3 & 2 & \\ 
Mistral Large 2 & 8.25 & 10 & 8 & 9 & 12 & 8 & 7 & 7 & 5 & \\

 \midrule 
 \textbf{Turn 2} & \textbf{Avg.} & \textbf{English} & \textbf{French} & \textbf{Russian} & \textbf{Hindi} & \textbf{Italian} & \textbf{Portuguese} & \textbf{Spanish} & \textbf{Chinese} \\ \midrule 
o1-preview & 1.75 & 2 & 2 & 4 & 1 & 1 & 1 & 2 & 1 & \\ 
o1-mini & 2.50 & 4 & 3 & 2 & 2 & 2 & 3 & 1 & 3 & \\ 
GPT-4o & 5.50 & 5 & 7 & 9 & 4 & 5 & 5 & 5 & 4 & \\ 
GPT-4 & 7.25 & 8 & 6 & 7 & 8 & 7 & 7 & 8 & 7 & \\ 
Llama 3.1 405B & 2.00 & 1 & 1 & 1 & 3 & 3 & 2 & 3 & 2 & \\ 
Llama 3.1 70B & 4.50 & 3 & 4 & 6 & 5 & 4 & 4 & 4 & 6 & \\ 
Llama 3.1 8B & 12.12 & 11 & 11 & 13 & 13 & 12 & 12 & 11 & 14 & \\ 
Gemini-1.5 pro & 12.12 & 12 & 13 & 12 & 10 & 14 & 13 & 13 & 10 & \\ 
Gemini-1.5 flash & 13.62 & 14 & 14 & 14 & 14 & 13 & 14 & 14 & 12 & \\ 
Claude-3.5 Sonnet & 7.00 & 6 & 8 & 5 & 6 & 8 & 8 & 7 & 8 & \\ 
Claude-3-Sonnet & 9.25 & 9 & 10 & 8 & 7 & 10 & 9 & 10 & 11 & \\ 
Claude-3-Haiku & 11.75 & 13 & 12 & 11 & 11 & 11 & 11 & 12 & 13 & \\ 
Qwen-2.5 72B & 5.88 & 7 & 5 & 3 & 9 & 6 & 6 & 6 & 5 & \\ 
Mistral Large 2 & 9.75 & 10 & 9 & 10 & 12 & 9 & 10 & 9 & 9 & \\

 \midrule 
 \textbf{Turn 3} & \textbf{Avg.} & \textbf{English} & \textbf{French} & \textbf{Russian} & \textbf{Hindi} & \textbf{Italian} & \textbf{Portuguese} & \textbf{Spanish} & \textbf{Chinese} \\ \midrule 
o1-preview & 1.62 & 2 & 2 & 3 & 1 & 1 & 2 & 1 & 1 & \\ 
o1-mini & 3.50 & 4 & 4 & 2 & 3 & 4 & 3 & 3 & 5 & \\ 
GPT-4o & 5.38 & 6 & 5 & 7 & 5 & 6 & 5 & 6 & 3 & \\ 
GPT-4 & 7.50 & 8 & 6 & 9 & 8 & 8 & 7 & 7 & 7 & \\ 
Llama 3.1 405B & 1.50 & 1 & 1 & 1 & 2 & 2 & 1 & 2 & 2 & \\ 
Llama 3.1 70B & 4.12 & 3 & 3 & 6 & 4 & 3 & 4 & 4 & 6 & \\ 
Llama 3.1 8B & 11.38 & 11 & 11 & 13 & 12 & 10 & 10 & 10 & 14 & \\ 
Gemini-1.5 pro & 11.62 & 10 & 12 & 11 & 11 & 13 & 13 & 13 & 10 & \\ 
Gemini-1.5 flash & 13.62 & 14 & 14 & 14 & 13 & 14 & 14 & 14 & 12 & \\ 
Claude-3.5 Sonnet & 5.75 & 5 & 7 & 4 & 6 & 7 & 8 & 5 & 4 & \\ 
Claude-3-Sonnet & 8.62 & 7 & 9 & 8 & 7 & 9 & 9 & 9 & 11 & \\ 
Claude-3-Haiku & 11.75 & 13 & 13 & 10 & 10 & 12 & 11 & 12 & 13 & \\ 
Qwen-2.5 72B & 7.25 & 9 & 8 & 5 & 9 & 5 & 6 & 8 & 8 & \\ 
Mistral Large 2 & 11.38 & 12 & 10 & 12 & 14 & 11 & 12 & 11 & 9 & \\

    \bottomrule
    \end{tabularx}
    \end{table}

\section{Experiments}
In this section, we present our methodology for benchmarking state-of-the-art LLMs on the \dataset benchmark, detailing our evaluation protocol and metrics. We then discuss the overall results across 14 leading models, highlighting key performance trends and cross-lingual variations in multi-turn instruction following tasks.


\subsection{Evaluation Protocol and Metrics} 
Here is the process of how we obtain LLM responses for the multi-turn conversations: For the first turn, the prompt of this turn is presented to the LLM, and its corresponding response is collected. For each successive turn, the preceding turns' prompts and responses are concatenated with the current turn's prompt, forming a cumulative input string. This concatenated string is then utilized to generate a response for the current turn via the LLM.

We report the following four accuracy scores:
\begin{itemize}
    \item Instruction-level strict accuracy: This metric measures the percentage of individual instructions that are accurately followed by LLMs.  It assesses the precision in executing each specific instruction given by the user.
    \item Conversation-level strict accuracy: This metric evaluates the accuracy across an entire multi-turn conversation, which may contain multiple instructions. It measures the percentage of conversations whose every instruction, \textit{from the first user turn to the current turn}, is followed correctly by LLMs. This ensures that the AI consistently understands and executes all directives throughout a conversation.
    \item Instruction-level loose accuracy: This is the instruction-level accuracy computed with a loose criterion following ~\citep{zhou2023instruction}. For example, responses of a LLM may start with an introductory sentence like "Certainly, here is the revised version..." and this sentence often violate instruction types like "All Uppercase". To resolve this issue, the loose criterion will remove the first sentence and then perform the evaluation. 
    \item Conversation-level loose accuracy: Likewise, conversation-level accuracy computed with a loose criterion.
\end{itemize}
We report the average of the above four accuracy scores as the \textbf{final metric} for each model on each turn and each language. More details about evaluation can be found in Appendix~\ref{app:exp-details}.

\subsection{Overall Results}
We benchmark 14 models: OpenAI o1-preview, o1-mini~\citep{openai2024o1}, GPT-4 (gpt-4-turbo-2024-04-09), GPT-4o~\citep{achiam2023gpt}, Llama 3.1 (8B, 70B, and 405B)~\citep{dubey2024llama}, Gemini-1.5 Pro (gemini-1.5-pro-preview-0514), Gemini-1.5 Flash (gemini-1.5-flash-preview-0514)~\citep{reid2024gemini}, Claude-3.5-Sonnet, Claude-3-Sonnet, Claude-3-Haiku~\citep{TheC3}, Qwen-2.5 72B~\citep{qwen2.5}, and Mistral Large 2~\citep{mistral2024large}, using the \dataset benchmark.


A summary of the overall results is presented in Figure~\ref{fig:overall_results}. In terms of the average score across all languages and across all turns. o1-preview, Llama 3.1 405B and o1-mini achieves the top 3 performance.

The evaluation results of each turn and each language are summarized in Table~\ref{tab:evaluation-results-accuracy-main} and Table~\ref{tab:evaluation-results-ranks-main} for detailed accuracy and ranking scores. For further analysis, we focused on the models’ average accuracy across all languages in the final turn, due to the multi-turn nature of the benchmark. OpenAI o1-preview and LLama 3.1 405B achieved the highest accuracy score of 0.707, closely followed by o1-mini with an average accuracy of 0.681. The Llama 3.1 70B model is scored 0.668, ranked the fourth. When comparing o1-preview and Llama 3.1 405B, Llama 3.1 405B significantly outperforms o1-preview on Russian, whereas o1-preview excels in Hindi, Chinese and English. For other languages, both models perform comparably. 

Other models, such as GPT-4o, GPT-4, Claude-3.5 Sonnet and Qwen-2.5 72B, also performed similarly, all achieving an average accuracy above 0.6. Models scoring below 0.6 by the third turn include several others, which are mainly due to either 1) poor performance on the first turn; or 2) a more significant drop in performance between consecutive turns. Notably, Qwen-2.5 72B, ranked $4_{\text{th}}$ in the first turn, saw a significant drop in accuracy by the third turn (ranked around $7_{\text{th}}$). This indicates that certain models struggle with the increasing complexity of multi-turn conversations and may not generalize well across multiple turns. 

Nearly all models showed worse performance on Russian compared to other languages, likely due to insufficient training data. Mistral Large, for instance, performed well on languages commonly spoken in Europe—English, French, Italian, Portuguese, and Spanish—but its poor performance on Russian, Hindi, and Chinese lowered its overall ranking. Additionally, for Llama 3.1 models (8B, 70B, and 405B), we observe a significant improvement in instruction-following accuracy as model size increases, particularly across the three turns, underscoring the benefits of scaling up models. Lastly, in our experiments with Google's Gemini models, we found that Gemini tends to be overly conservative, even refusing reasonable user requests like capitalizing responses. However, this isn't the primary reason for the weaker performance of Gemini 1.5. We suspect the main issue is that the Gemini models (e.g., gemini-1.5-prop-0514) simply aren't as performant as others. This is evident from its first-turn performance on the English split, which is consistent with observations on Scale AI's leaderboard.\footnote{\url{https://scale.com/leaderboard/instruction_following}}We are aware of the latest version of Gemini (gemini-1.5-pro-exp-0827) but it is in experimental mode and hardly conduct large-scale test of it.

\begin{figure}[t!]
  \centering
  \setlength{\abovecaptionskip}{0.0cm}
  \setlength{\belowcaptionskip}{0.0cm}
  \includegraphics[width=0.9\linewidth]{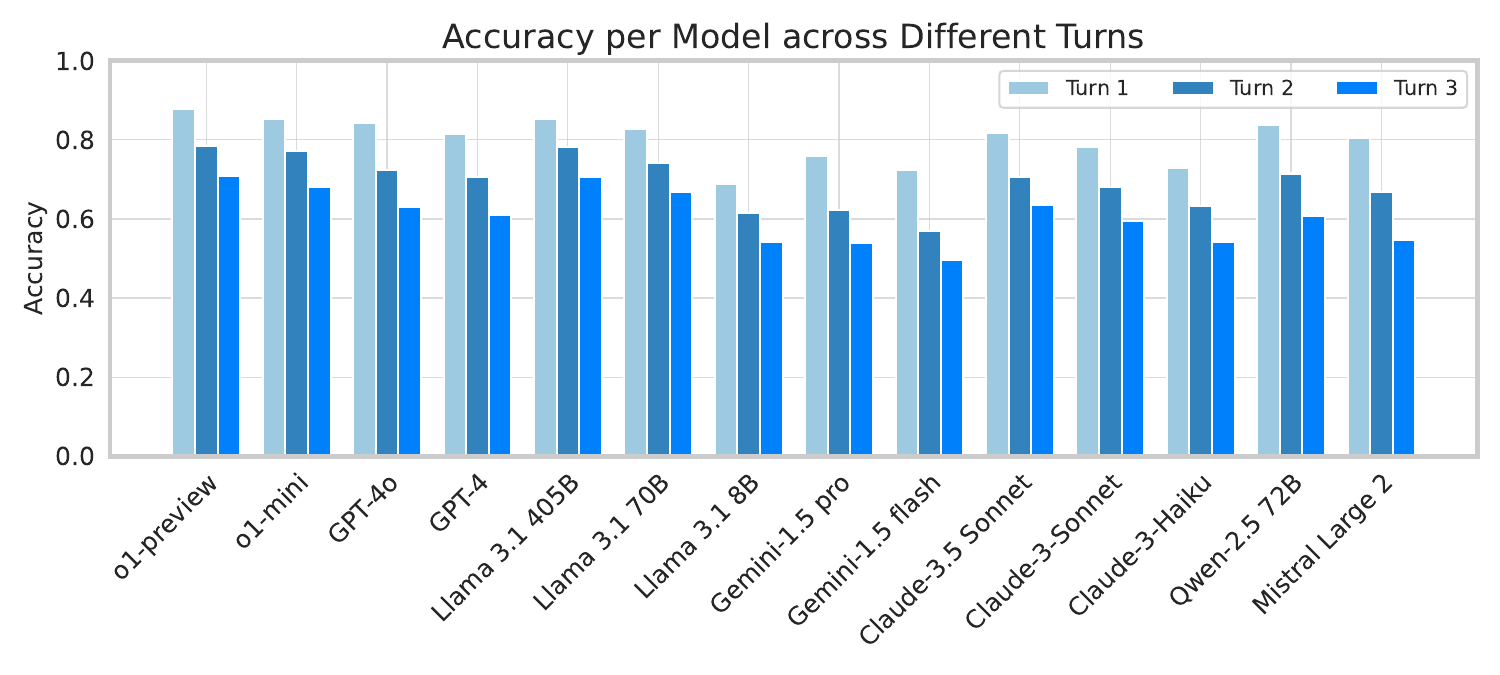}
  \caption{The impact of multi-turns on instruction following. ``Accuracy" means the average of the final metric of the all languages; The final metric is the average of the four accuracy scores: instruction-level strict accuracy, conversation-level strict accuracy, instruction-level loose accuracy, and conversation-level loose accuracy. }
\label{fig: impact of multi-turns}
\end{figure}
\section{Analysis}
In this section, we provide a detailed understanding of the evaluation results and delve into various aspects of language model performance on instruction following tasks. We analyze the impact of multi-turn conversations on instruction following accuracy, explore the phenomenon of instruction forgetting across turns, and examine models' ability to recover from errors in previous turns. Additionally, we investigate the performance of language models across different languages and conduct an in-depth error analysis to identify patterns and challenges in multilingual instruction following. These analyses offer valuable insights into the strengths and limitations of current language models in complex, multi-turn, and multilingual scenarios.

\subsection{Multi-Turn Instruction Following}

The visualization of the impact from multi-turn instructions is presented in Figure \ref{fig: impact of multi-turns}, where ``Accuracy'' refers to the average final metric across all languages. The final metric is the average of instruction-level strict accuracy, conversation-level strict accuracy, instruction-level loose accuracy, and conversation-level loose accuracy. All models struggle increasingly with following multi-turn instructions as the number of turns increases. All the models tested showed a higher rate of failure in executing instructions correctly with each additional turn. For example, o1-preview drops from 0.877 at the first turn to 0.707 at the third turn in terms of average accuracy over all languages. Additionally, different models exhibit different patterns across three turns. Some models, such as Qwen 2.5 72B Instruction~\citep{qwen2.5}, may achieve a high accuracy than other models, but degrades much faster than the others in later turns. In contrast, models such as o1-preview, o1-mini and Llama 3.1 405B instruct, suffers less degradation in accuracy after multiple turns of interactions. In the following sections, we seek to provide a detailed analysis to understanding such differences.

\subsection{Instructions Forgetting to Follow in Multi-Turns}

\begin{figure}[t!]
    \centering
    \includegraphics[width=0.9\linewidth]{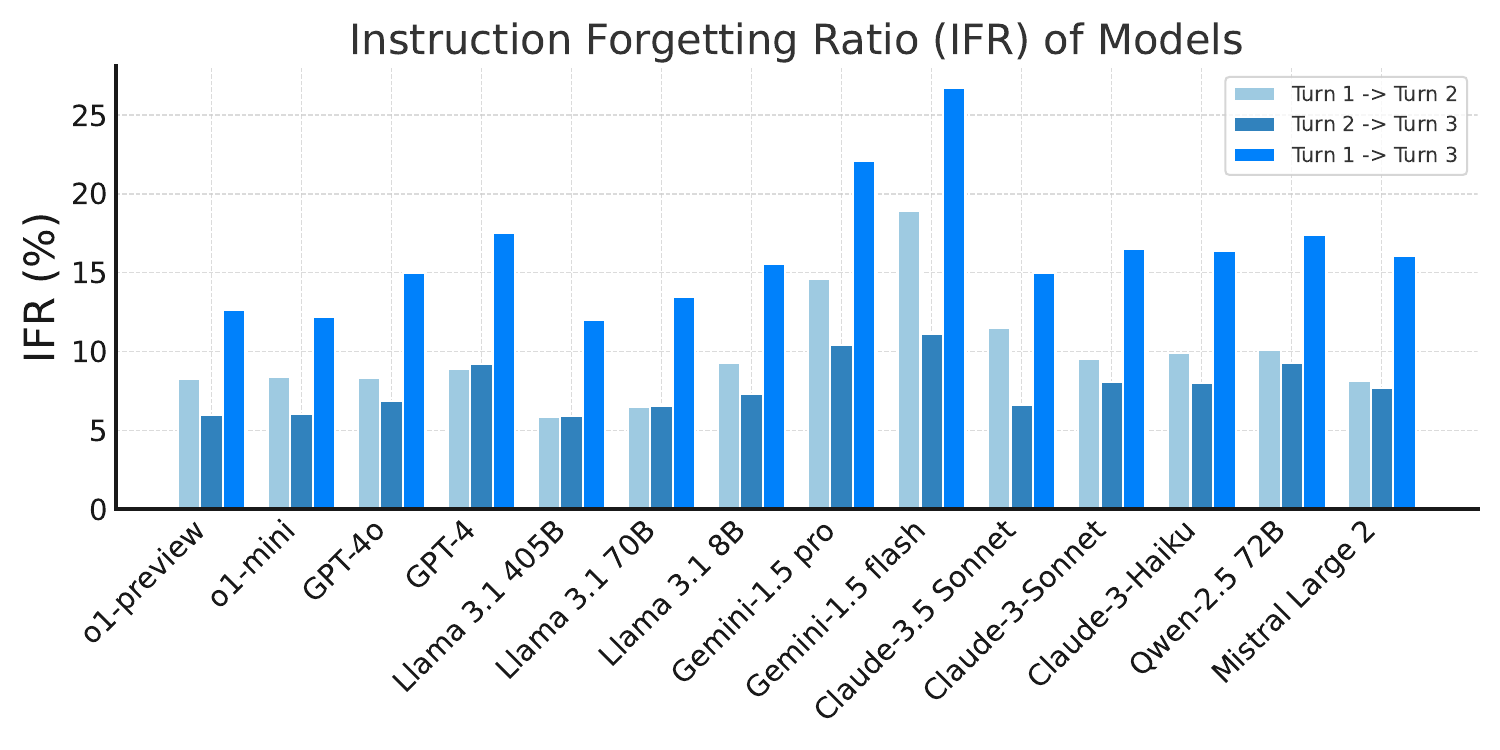}
    \caption{Instruction Forgetting Ratio (IFR) of Models: This figure shows the IFR for various models, highlighting their ability to retain instruction information across multiple turns.}
    \label{fig:ifr-plot}
\end{figure}

In this study, we investigate the factors contributing to the decrease in accuracy of LLMs when following instructions in multi-turn conversations. Specifically, we aim to determine whether the drop in accuracy is primarily due to the multiplication of single-turn error rate or if it is caused by the increased difficulty resulting from more rounds and instructions being combined. Our pilot study focuses on the forgetting issue that arises throughout the multi-turn conversation.

We first define Instruction Forgetting Ratio (IFR): The percentage of previously followed instructions that were not followed in the subsequent turn, defined as below:
\begin{equation}
\text{IFR} = \left( \frac{\text{Number of Previously Followed Instructions Not Followed in Subsequent Turns}}{\text{Total Number of Instructions Followed in Previous Turn}} \right) \times 100
\end{equation}

For example, the instruction of ``You are not allowed to use any commas in your response'' in turn 1 in Figure \ref{fig: multiif_method} could be followed by the response to turn 1 but then not followed by that of turn 2. We count this case as the instruction that is previously followed by not followed in subsequent turns.

Using this definition, we can calculate the IFR of each turn's instructions in subsequent turns. The IFR values are visualized in a bar plot presented in Figure \ref{fig:ifr-plot}. Our Analysis of the visualization reveals a pronounced tendency for LLMs to exhibit decreased adherence to previously executed instructions as the number of turns progresses. Specifically, for high-performing models, such as o1-preview, o1-mini, Llama 3.1 (70B and 405B), and Claude 3.5 Sonnet, generally have a lower rate of instruction forgetting. In particular, for Llama 3.1 models, the forgetting rate (i.e., IFR) also decreases as the model size is scaled up from 8B to 405B. Gemini models (Gemini-1.5 pro and Gemini-1.5 flash) exhibit the highest rate of instruction forgetting, which is partly but not mainly due to the false refusals in responses from Gemini models in our later investigations (check Appendix~\ref{app:gemini-false-refusals} for more details). Lastly and surprisingly, the IFR rates from turn 1 to turn 2 are generally higher than the IFR rate from turn 2 to turn 3. Further investigation is warranted to elucidate the underlying causes of this forgetting phenomenon, and targeted strategies will be developed to mitigate it.

\subsection{Error Self-Correction in Multi-Turns}\label{subsec:ecr}
\begin{figure}[t!]
    \centering
    \includegraphics[width=0.9\linewidth]{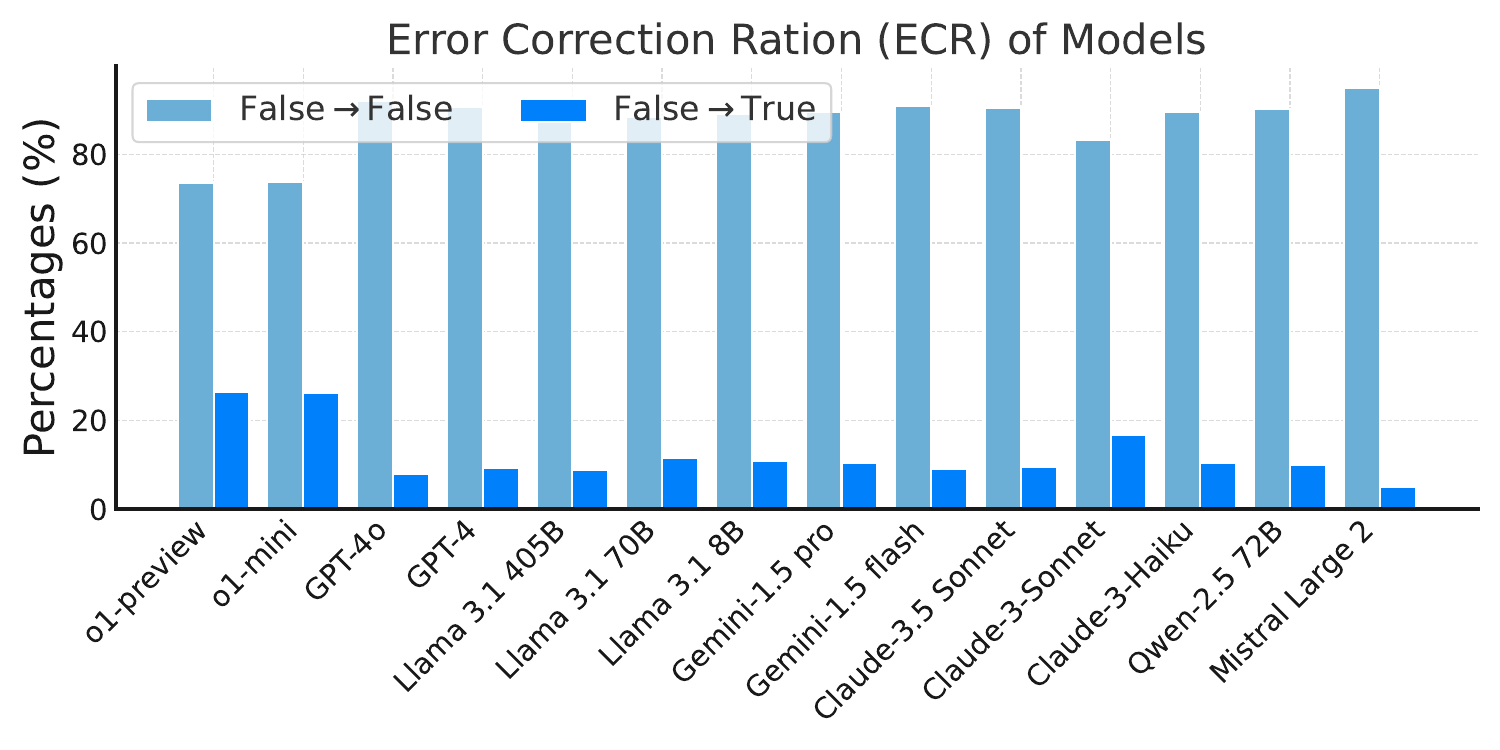}
    \caption{Error Correction Ratio (ECR) of Models (i.e., False $\rightarrow$ True): This figure shows the ECR for various models, as well as the portion of responses (i.e., the light blue bars) that did not correct from errors in prior turns. }
    \label{fig:error-recovery}
\end{figure}
Instruction forgetting is a key issue contributing to degraded performance in multi-turn instruction following. In this section, we explore whether the model can recover or recall forgotten instructions in later turns. The ability to recover from errors made in previous turns is a crucial capability for advanced LLMs, especially in complex environments requiring multi-step actions, such as those involving LLM agents. To assess this, we compute the error correction rate using the following equation:
\begin{equation}
\text{ECR} = \left( \frac{\text{Number of Previously \textbf{Unfollowed} Instructions Followed in Subsequent Turns}}{\text{Total Number of Instructions \textbf{UnFollowed} in Previous Turn}} \right) \times 100
\end{equation}
The results, shown in Figure~\ref{fig:error-recovery}, present the Error Correction Ratio (ECR). OpenAI's o1-preview and o1-mini models exhibit the highest ECR—correcting around $25\%$ of unfollowed instructions in later turns—suggesting that their incorporation of a hidden chain of thought is especially helpful in error correction. In contrast, other models perform similarly in terms of ECR. This is likely because these models do not "reflect" on their outputs to fix prior errors, as doing so would violate the instruction-following criteria. Thus, the hidden chain of thought capability plays a key role in error correction during multi-turn interactions. This also highlights opportunities for improvement in models that do not yet incorporate this feature, though all the models exhibit some levels of error recovery from errors made in prior turns.

\begin{figure}[t!]
  \centering
  \setlength{\abovecaptionskip}{0.0cm}
  \setlength{\belowcaptionskip}{0.0cm}
\includegraphics[width=0.7\linewidth]{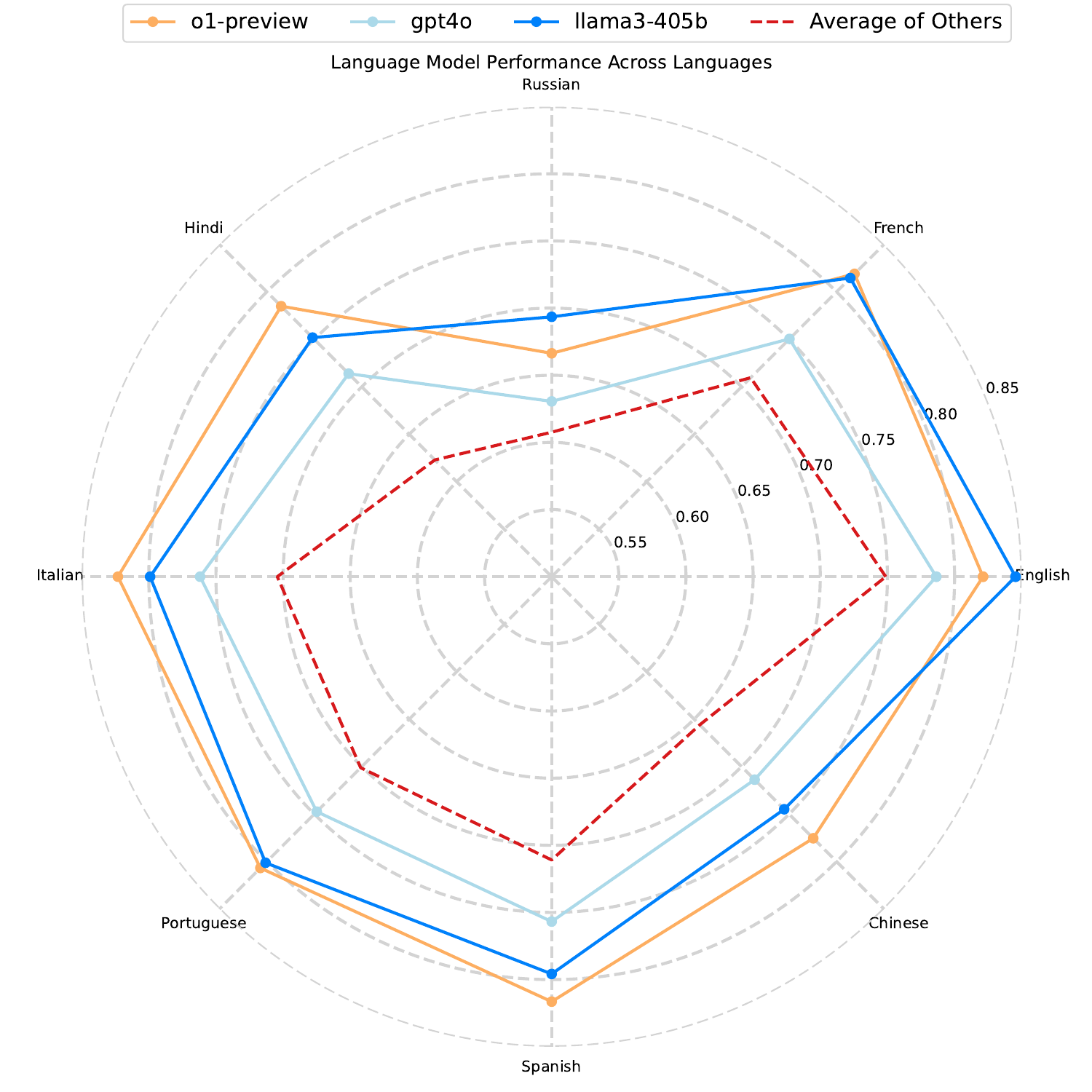}
  \caption{The performance of multilingual instructions. The performance is measured by the average of the four accuracy scores across all three turns: instruction-level strict accuracy, conversation-level strict accuracy, instruction-level loose accuracy, and conversation-level loose accuracy.}
\label{fig:multilingual_impact_figure}
\end{figure}

\subsection{Multilingual Instruction Following}

The performance of three language models (o1-preview, gpt4o, and llama3-405b) across eight languages is illustrated in Figure \ref{fig:multilingual_impact_figure}. We observe that English has the best scores for all models, reaching approximately 0.85 with Llama 3.1 405B, followed closely by French and Italian.  The o1-preview model generally achieves the highest performance across languages, with Llama 3.1 405B following closely behind. There is notable variation in performance for languages like Russian and Hindi. The ``Average of Others" curve demonstrates significantly lower performance across all languages compared to the three highlighted models, suggesting a gap between these advanced models and other models. 

Interestingly, o1-preview and llama 3.1 405B show comparable performance in most languages, with o1-preview having a slight edge in languages like Chinese, Spanish, Italian, and Hindi. The GPT-4o model, while performing well, often falls slightly behind the other two models. This comparison reveals the current state of multilingual capabilities in advanced language models, with o1-preview being the best-performing model. It also highlights areas for potential improvement, such as on languages like Russian, Hindi, and Chinese.

\subsection{Error Analysis of Multilingual Instruction Following}

\begin{figure}[h]
    \centering
    \includegraphics[width=\linewidth]{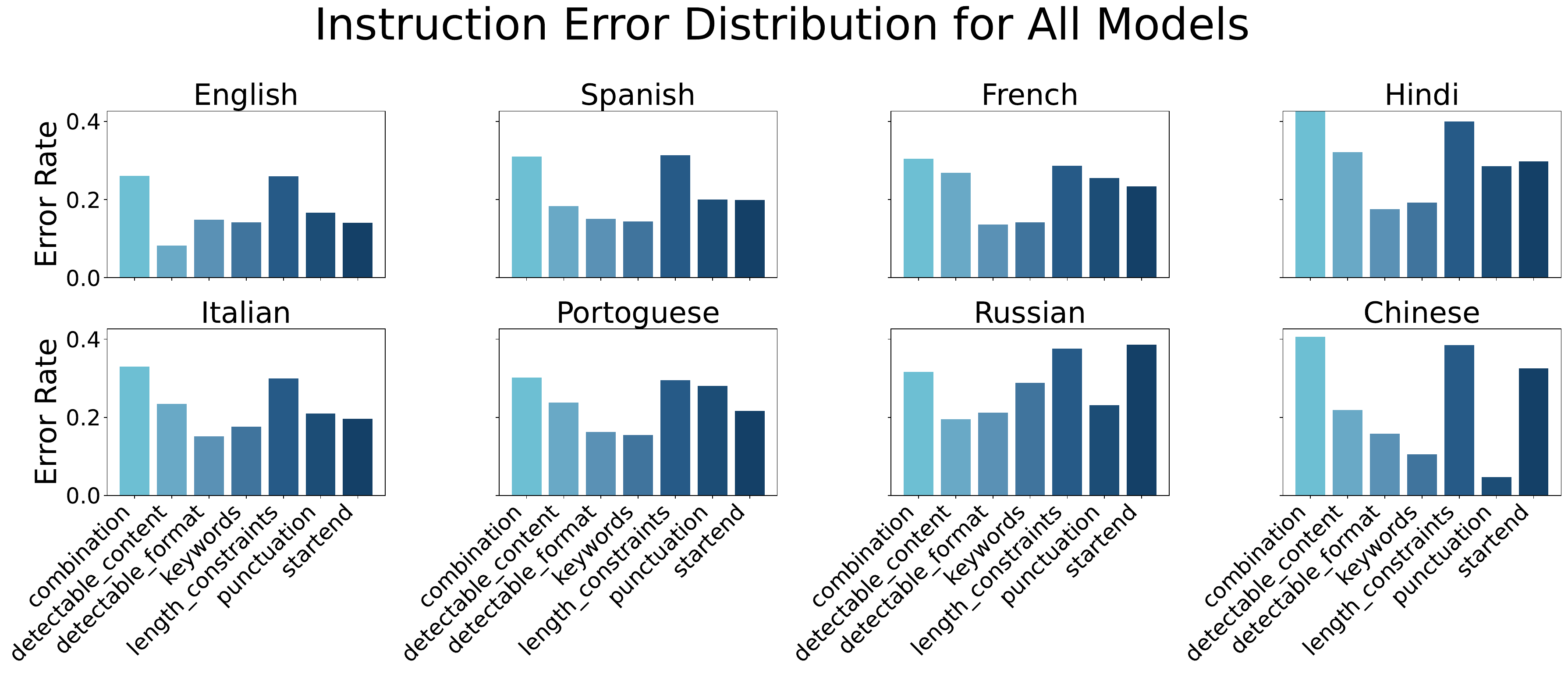}
    \caption{Instruction Error Distribution Across Languages and Instruction Categories for Aggregated Model Performance. The chart displays error rates for various instruction types in eight languages, highlighting cross-lingual patterns and category-specific challenges in language model instruction following. 
    }
    \label{fig:instruction_error_distribution}
\end{figure}

Figure \ref{fig:instruction_error_distribution} presents the error distribution for various instruction categories across eight languages, aggregating results from all models tested. We also provide the error distribution for each evaluated model in the Appendix~\ref{app:error_distribution}. This analysis reveals significant variations in error rates across both languages and instruction types. Languages with non-Latin scripts (Hindi, Russian, and Chinese) generally exhibit higher error rates, suggesting potential limitations in the models' multilingual capabilities. English consistently shows lower error rates, possibly due to its prevalence in training data.

Certain instruction categories prove challenging across languages. The \textit{length\_constraint} and \textit{combination} categories show consistently high error rates, indicating universal difficulties in precise output length control and text structure control. The \textit{startend} category displays high variability, with particularly elevated error rates in Russian and Chinese. 

Language-specific observations include notably high error rates for Hindi across most categories, high variability in Russian (especially for \textit{startend}), and a distinct pattern in Chinese with elevated errors in \textit{combination}, \textit{length\_constraint}, and \textit{startend} categories.
These findings highlight areas for improvement in multilingual instruction-following capabilities, and it also emphasizes the need for targeted enhancements in consistently challenging instruction types and better support for non-Latin script languages. Future research should focus on addressing these performance gaps and investigating the factors contributing to relative successes in categories like \textit{detectcable\_format}.


\section{Conclusion}
In this study, we introduced \dataset, a novel benchmark designed to evaluate the proficiency of LLMs in multi-turn and multilingual instruction following. Existing benchmarks like IFEval provide a solid foundation for assessing instruction adherence in single-turn scenarios, however, they fall short in capturing the complexities of real-world interactions that often involve multiple turns and multiple languages. The results from \dataset indicate that current leading LLMs, including o1-preview, o1-mini, Llama 3.1 and GPT4-o, struggle with maintaining high accuracy in following multi-turn instructions and exhibit even lower performance in non-English languages.

\bibliographystyle{assets/plainnat}
\bibliography{paper}

\begin{thebibliography}{43}
\providecommand{\natexlab}[1]{#1}
\providecommand{\url}[1]{\texttt{#1}}
\expandafter\ifx\csname urlstyle\endcsname\relax
  \providecommand{\doi}[1]{doi: #1}\else
  \providecommand{\doi}{doi: \begingroup \urlstyle{rm}\Url}\fi

\bibitem[Achiam et~al.(2023)Achiam, Adler, Agarwal, Ahmad, Akkaya, Aleman,
  Almeida, Altenschmidt, Altman, Anadkat, et~al.]{achiam2023gpt}
Josh Achiam, Steven Adler, Sandhini Agarwal, Lama Ahmad, Ilge Akkaya,
  Florencia~Leoni Aleman, Diogo Almeida, Janko Altenschmidt, Sam Altman,
  Shyamal Anadkat, et~al.
\newblock Gpt-4 technical report.
\newblock \emph{arXiv preprint arXiv:2303.08774}, 2023.

\bibitem[Anthropic()]{TheC3}
Anthropic.
\newblock The claude 3 model family: Opus, sonnet, haiku.
\newblock \url{https://api.semanticscholar.org/CorpusID:268232499}.

\bibitem[Blasi et~al.(2021)Blasi, Anastasopoulos, and
  Neubig]{blasi2021systematic}
Damian Blasi, Antonios Anastasopoulos, and Graham Neubig.
\newblock Systematic inequalities in language technology performance across the
  world's languages.
\newblock \emph{arXiv preprint arXiv:2110.06733}, 2021.

\bibitem[Budzianowski et~al.(2018)Budzianowski, Wen, Tseng, Casanueva, Ultes,
  Ramadan, and Ga{\v{s}}i{\'c}]{budzianowski2018multiwoz}
Pawe{\l} Budzianowski, Tsung-Hsien Wen, Bo-Hsiang Tseng, Inigo Casanueva,
  Stefan Ultes, Osman Ramadan, and Milica Ga{\v{s}}i{\'c}.
\newblock Multiwoz--a large-scale multi-domain wizard-of-oz dataset for
  task-oriented dialogue modelling.
\newblock \emph{arXiv preprint arXiv:1810.00278}, 2018.

\bibitem[Chang et~al.(2024)Chang, Wang, Wang, Wu, Yang, Zhu, Chen, Yi, Wang,
  Wang, et~al.]{chang2024survey}
Yupeng Chang, Xu~Wang, Jindong Wang, Yuan Wu, Linyi Yang, Kaijie Zhu, Hao Chen,
  Xiaoyuan Yi, Cunxiang Wang, Yidong Wang, et~al.
\newblock A survey on evaluation of large language models.
\newblock \emph{ACM Transactions on Intelligent Systems and Technology},
  15\penalty0 (3):\penalty0 1--45, 2024.

\bibitem[Chen et~al.(2024)Chen, Du, Wen, Zhou, Cui, Weng, Tu, Wang, Tong,
  Huang, et~al.]{chen2024mj}
Zhaorun Chen, Yichao Du, Zichen Wen, Yiyang Zhou, Chenhang Cui, Zhenzhen Weng,
  Haoqin Tu, Chaoqi Wang, Zhengwei Tong, Qinglan Huang, et~al.
\newblock Mj-bench: Is your multimodal reward model really a good judge for
  text-to-image generation?
\newblock \emph{arXiv preprint arXiv:2407.04842}, 2024.

\bibitem[Crook and Marin(2017)]{asri2016sequence}
Paul~A Crook and Alex Marin.
\newblock Sequence to sequence modeling for user simulation in dialog systems.
\newblock In \emph{Interspeech}, pages 1706--1710, 2017.

\bibitem[Dubey et~al.(2024)Dubey, Jauhri, Pandey, Kadian, Al-Dahle, Letman,
  Mathur, Schelten, Yang, Fan, et~al.]{dubey2024llama}
Abhimanyu Dubey, Abhinav Jauhri, Abhinav Pandey, Abhishek Kadian, Ahmad
  Al-Dahle, Aiesha Letman, Akhil Mathur, Alan Schelten, Amy Yang, Angela Fan,
  et~al.
\newblock The llama 3 herd of models.
\newblock \emph{arXiv preprint arXiv:2407.21783}, 2024.

\bibitem[He et~al.(2024)He, Zeng, Huang, Chen, Xiao, He, Zhou, Liang, and
  Xiao]{he2024can}
Qianyu He, Jie Zeng, Wenhao Huang, Lina Chen, Jin Xiao, Qianxi He, Xunzhe Zhou,
  Jiaqing Liang, and Yanghua Xiao.
\newblock Can large language models understand real-world complex instructions?
\newblock In \emph{Proceedings of the AAAI Conference on Artificial
  Intelligence}, volume~38, pages 18188--18196, 2024.

\bibitem[Jiang et~al.(2023)Jiang, Wang, Zeng, Zhong, Li, Mi, Shang, Jiang, Liu,
  and Wang]{jiang2023followbench}
Yuxin Jiang, Yufei Wang, Xingshan Zeng, Wanjun Zhong, Liangyou Li, Fei Mi,
  Lifeng Shang, Xin Jiang, Qun Liu, and Wei Wang.
\newblock Followbench: A multi-level fine-grained constraints following
  benchmark for large language models.
\newblock \emph{arXiv preprint arXiv:2310.20410}, 2023.

\bibitem[Jin et~al.(2023)Jin, Mehri, Hazarika, Padmakumar, Lee, Liu, and
  Namazifar]{jin2023data}
Di~Jin, Shikib Mehri, Devamanyu Hazarika, Aishwarya Padmakumar, Sungjin Lee,
  Yang Liu, and Mahdi Namazifar.
\newblock Data-efficient alignment of large language models with human feedback
  through natural language.
\newblock \emph{arXiv preprint arXiv:2311.14543}, 2023.

\bibitem[Kocmi and Federmann(2023)]{kocmi2023large}
Tom Kocmi and Christian Federmann.
\newblock Large language models are state-of-the-art evaluators of translation
  quality.
\newblock \emph{arXiv preprint arXiv:2302.14520}, 2023.

\bibitem[Li et~al.(2023)Li, Zhang, Dubois, Taori, Gulrajani, Guestrin, Liang,
  and Hashimoto]{li2023alpacaeval}
Xuechen Li, Tianyi Zhang, Yann Dubois, Rohan Taori, Ishaan Gulrajani, Carlos
  Guestrin, Percy Liang, and Tatsunori~B Hashimoto.
\newblock Alpacaeval: An automatic evaluator of instruction-following models,
  2023.

\bibitem[Li et~al.(2017)Li, Su, Shen, Li, Cao, and Niu]{li2017dailydialog}
Yanran Li, Hui Su, Xiaoyu Shen, Wenjie Li, Ziqiang Cao, and Shuzi Niu.
\newblock Dailydialog: A manually labelled multi-turn dialogue dataset.
\newblock \emph{arXiv preprint arXiv:1710.03957}, 2017.

\bibitem[Lin(2004)]{lin2004rouge}
Chin-Yew Lin.
\newblock Rouge: A package for automatic evaluation of summaries.
\newblock In \emph{Text summarization branches out}, pages 74--81, 2004.

\bibitem[Liu et~al.(2016)Liu, Lowe, Serban, Noseworthy, Charlin, and
  Pineau]{liu2016not}
Chia-Wei Liu, Ryan Lowe, Iulian~V Serban, Michael Noseworthy, Laurent Charlin,
  and Joelle Pineau.
\newblock How not to evaluate your dialogue system: An empirical study of
  unsupervised evaluation metrics for dialogue response generation.
\newblock \emph{arXiv preprint arXiv:1603.08023}, 2016.

\bibitem[Liu et~al.(2023{\natexlab{a}})Liu, Lei, Wang, Huang, Feng, Wen, Cheng,
  Ke, Xu, Tam, et~al.]{liu2023alignbench}
Xiao Liu, Xuanyu Lei, Shengyuan Wang, Yue Huang, Zhuoer Feng, Bosi Wen, Jiale
  Cheng, Pei Ke, Yifan Xu, Weng~Lam Tam, et~al.
\newblock Alignbench: Benchmarking chinese alignment of large language models.
\newblock \emph{arXiv preprint arXiv:2311.18743}, 2023{\natexlab{a}}.

\bibitem[Liu et~al.(2023{\natexlab{b}})Liu, Iter, Xu, Wang, Xu, and
  Zhu]{liu2023g}
Yang Liu, Dan Iter, Yichong Xu, Shuohang Wang, Ruochen Xu, and Chenguang Zhu.
\newblock G-eval: Nlg evaluation using gpt-4 with better human alignment.
\newblock \emph{arXiv preprint arXiv:2303.16634}, 2023{\natexlab{b}}.

\bibitem[Mehri and Eskenazi(2020)]{mehri2020usr}
Shikib Mehri and Maxine Eskenazi.
\newblock Usr: An unsupervised and reference free evaluation metric for dialog
  generation.
\newblock \emph{arXiv preprint arXiv:2005.00456}, 2020.

\bibitem[{Mistral AI team}(2024)]{mistral2024large}
{Mistral AI team}.
\newblock Mistral large 2, 2024.
\newblock \url{https://mistral.ai/news/mistral-large-2407/}.

\bibitem[OpenAI()]{openaio1}
OpenAI.
\newblock "https://openai.com/index/introducing-openai-o1-preview/".

\bibitem[{OpenAI}(2024)]{openai2024o1}
{OpenAI}.
\newblock Openai o1 system card, September 2024.
\newblock \url{https://openai.com/index/}.

\bibitem[Ouyang et~al.(2022)Ouyang, Wu, Jiang, Almeida, Wainwright, Mishkin,
  Zhang, Agarwal, Slama, Ray, et~al.]{ouyang2022training}
Long Ouyang, Jeffrey Wu, Xu~Jiang, Diogo Almeida, Carroll Wainwright, Pamela
  Mishkin, Chong Zhang, Sandhini Agarwal, Katarina Slama, Alex Ray, et~al.
\newblock Training language models to follow instructions with human feedback.
\newblock \emph{Advances in neural information processing systems},
  35:\penalty0 27730--27744, 2022.

\bibitem[Papineni et~al.(2002)Papineni, Roukos, Ward, and
  Zhu]{papineni2002bleu}
Kishore Papineni, Salim Roukos, Todd Ward, and Wei-Jing Zhu.
\newblock Bleu: a method for automatic evaluation of machine translation.
\newblock In \emph{Proceedings of the 40th annual meeting of the Association
  for Computational Linguistics}, pages 311--318, 2002.

\bibitem[Qin et~al.(2024)Qin, Song, Hu, Yao, Cho, Wang, Wu, Liu, Liu, and
  Yu]{qin2024infobench}
Yiwei Qin, Kaiqiang Song, Yebowen Hu, Wenlin Yao, Sangwoo Cho, Xiaoyang Wang,
  Xuansheng Wu, Fei Liu, Pengfei Liu, and Dong Yu.
\newblock Infobench: Evaluating instruction following ability in large language
  models.
\newblock \emph{arXiv preprint arXiv:2401.03601}, 2024.

\bibitem[Reid et~al.(2024)Reid, Savinov, Teplyashin, Lepikhin, Lillicrap,
  Alayrac, Soricut, Lazaridou, Firat, Schrittwieser, et~al.]{reid2024gemini}
Machel Reid, Nikolay Savinov, Denis Teplyashin, Dmitry Lepikhin, Timothy
  Lillicrap, Jean-baptiste Alayrac, Radu Soricut, Angeliki Lazaridou, Orhan
  Firat, Julian Schrittwieser, et~al.
\newblock Gemini 1.5: Unlocking multimodal understanding across millions of
  tokens of context.
\newblock \emph{arXiv preprint arXiv:2403.05530}, 2024.

\bibitem[Ruder et~al.(2021)Ruder, Constant, Botha, Siddhant, Firat, Fu, Liu,
  Hu, Garrette, Neubig, et~al.]{ruder2021xtreme}
Sebastian Ruder, Noah Constant, Jan Botha, Aditya Siddhant, Orhan Firat, Jinlan
  Fu, Pengfei Liu, Junjie Hu, Dan Garrette, Graham Neubig, et~al.
\newblock Xtreme-r: Towards more challenging and nuanced multilingual
  evaluation.
\newblock \emph{arXiv preprint arXiv:2104.07412}, 2021.

\bibitem[Sellam et~al.(2020)Sellam, Das, and Parikh]{sellam2020bleurt}
Thibault Sellam, Dipanjan Das, and Ankur~P Parikh.
\newblock Bleurt: Learning robust metrics for text generation.
\newblock \emph{arXiv preprint arXiv:2004.04696}, 2020.

\bibitem[Siddhant et~al.(2020)Siddhant, Hu, Johnson, Firat, and
  Ruder]{siddhant2020xtreme}
Aditya Siddhant, Junjie Hu, Melvin Johnson, Orhan Firat, and Sebastian Ruder.
\newblock Xtreme: A massively multilingual multi-task benchmark for evaluating
  cross-lingual generalization.
\newblock In \emph{Proceedings of the International Conference on Machine
  Learning 2020}, pages 4411--4421, 2020.

\bibitem[Sun et~al.(2023)Sun, Tian, Zhou, Xu, Hu, Gupta, Wieting, Peng, and
  Ma]{sun2023evaluating}
Jiao Sun, Yufei Tian, Wangchunshu Zhou, Nan Xu, Qian Hu, Rahul Gupta,
  John~Frederick Wieting, Nanyun Peng, and Xuezhe Ma.
\newblock Evaluating large language models on controlled generation tasks.
\newblock \emph{arXiv preprint arXiv:2310.14542}, 2023.

\bibitem[Taori et~al.(2023)Taori, Gulrajani, Zhang, Dubois, Li, Guestrin,
  Liang, and Hashimoto]{taori2023stanford}
Rohan Taori, Ishaan Gulrajani, Tianyi Zhang, Yann Dubois, Xuechen Li, Carlos
  Guestrin, Percy Liang, and Tatsunori~B Hashimoto.
\newblock Stanford alpaca: An instruction-following llama model, 2023.

\bibitem[Team(2024)]{qwen2.5}
Qwen Team.
\newblock Qwen2.5: A party of foundation models, September 2024.
\newblock \url{https://qwenlm.github.io/blog/qwen2.5/}.

\bibitem[Tseng et~al.(2021)Tseng, Dai, Kreyssig, and
  Byrne]{tseng2021transferable}
Bo-Hsiang Tseng, Yinpei Dai, Florian Kreyssig, and Bill Byrne.
\newblock Transferable dialogue systems and user simulators.
\newblock \emph{arXiv preprint arXiv:2107.11904}, 2021.

\bibitem[Wang et~al.(2024)Wang, Jiang, Yang, Liu, and Chen]{wang2023beyond}
Chaoqi Wang, Yibo Jiang, Chenghao Yang, Han Liu, and Yuxin Chen.
\newblock Beyond reverse kl: Generalizing direct preference optimization with
  diverse divergence constraints.
\newblock \emph{International Conference on Learning Representations}, 2024.

\bibitem[Wang et~al.(2023)Wang, Li, Chen, Cai, Zhu, Lin, Cao, Liu, Liu, and
  Sui]{wang2023large}
Peiyi Wang, Lei Li, Liang Chen, Zefan Cai, Dawei Zhu, Binghuai Lin, Yunbo Cao,
  Qi~Liu, Tianyu Liu, and Zhifang Sui.
\newblock Large language models are not fair evaluators.
\newblock \emph{arXiv preprint arXiv:2305.17926}, 2023.

\bibitem[Wen et~al.(2024)Wen, Ke, Gu, Wu, Huang, Zhou, Li, Hu, Gao, Xu,
  et~al.]{wen2024benchmarking}
Bosi Wen, Pei Ke, Xiaotao Gu, Lindong Wu, Hao Huang, Jinfeng Zhou, Wenchuang
  Li, Binxin Hu, Wendy Gao, Jiaxin Xu, et~al.
\newblock Benchmarking complex instruction-following with multiple constraints
  composition.
\newblock \emph{arXiv preprint arXiv:2407.03978}, 2024.

\bibitem[Wu and Dredze(2020)]{wu2020are}
Shijie Wu and Mark Dredze.
\newblock Are all languages created equal in multilingual bert?
\newblock \emph{arXiv preprint arXiv:2005.09093}, 2020.

\bibitem[Xia et~al.(2024)Xia, Xing, Du, Yang, Feng, Xu, Yin, and
  Xiong]{xia2024fofo}
Congying Xia, Chen Xing, Jiangshu Du, Xinyi Yang, Yihao Feng, Ran Xu, Wenpeng
  Yin, and Caiming Xiong.
\newblock Fofo: A benchmark to evaluate llms' format-following capability.
\newblock \emph{arXiv preprint arXiv:2402.18667}, 2024.

\bibitem[Xu et~al.(2023)Xu, Sun, Zheng, Geng, Zhao, Feng, Tao, and
  Jiang]{xu2023wizardlm}
Can Xu, Qingfeng Sun, Kai Zheng, Xiubo Geng, Pu~Zhao, Jiazhan Feng, Chongyang
  Tao, and Daxin Jiang.
\newblock Wizardlm: Empowering large language models to follow complex
  instructions.
\newblock \emph{arXiv preprint arXiv:2304.12244}, 2023.

\bibitem[Zhang et~al.(2019)Zhang, Kishore, Wu, Weinberger, and
  Artzi]{zhang2020bertscore}
Tianyi Zhang, Varsha Kishore, Felix Wu, Kilian~Q Weinberger, and Yoav Artzi.
\newblock Bertscore: Evaluating text generation with bert.
\newblock \emph{arXiv preprint arXiv:1904.09675}, 2019.

\bibitem[Zheng et~al.(2023)Zheng, Chiang, Sheng, Zhuang, Wu, Zhuang, Lin, Li,
  Li, Xing, et~al.]{zheng2023judging}
Lianmin Zheng, Wei-Lin Chiang, Ying Sheng, Siyuan Zhuang, Zhanghao Wu, Yonghao
  Zhuang, Zi~Lin, Zhuohan Li, Dacheng Li, Eric Xing, et~al.
\newblock Judging llm-as-a-judge with mt-bench and chatbot arena.
\newblock \emph{Advances in Neural Information Processing Systems},
  36:\penalty0 46595--46623, 2023.

\bibitem[Zheng et~al.(2024)Zheng, Chiang, Sheng, Zhuang, Wu, Zhuang, Lin, Li,
  Li, Xing, et~al.]{zheng2024judging}
Lianmin Zheng, Wei-Lin Chiang, Ying Sheng, Siyuan Zhuang, Zhanghao Wu, Yonghao
  Zhuang, Zi~Lin, Zhuohan Li, Dacheng Li, Eric Xing, et~al.
\newblock Judging llm-as-a-judge with mt-bench and chatbot arena.
\newblock \emph{Advances in Neural Information Processing Systems}, 36, 2024.

\bibitem[Zhou et~al.(2023)Zhou, Lu, Mishra, Brahma, Basu, Luan, Zhou, and
  Hou]{zhou2023instruction}
Jeffrey Zhou, Tianjian Lu, Swaroop Mishra, Siddhartha Brahma, Sujoy Basu,
  Yi~Luan, Denny Zhou, and Le~Hou.
\newblock Instruction-following evaluation for large language models.
\newblock \emph{arXiv preprint arXiv:2311.07911}, 2023.

\end{thebibliography}

\newpage
\beginappendix


\section{Experimental Details}\label{app:exp-details}
\begin{table}[ht]
    \centering
    \begin{tabular}{l l c c c}
        \toprule
        \textbf{Model Name} & \textbf{Version} & \textbf{Temperature} & \textbf{Top-p} & \textbf{Max New Tokens} \\
        \midrule
        o1-preview & o1-preview-2024-09-12 & 1.0 & 1.0 & 25000 \\
        o1-mini    & o1-mini-2024-09-12    & 1.0 & 1.0 & 25000 \\
        GPT-4 & gpt-4-turbo-2024-04-09 & 0.6 & 0.9 & 1024 \\
        GPT-4o & gpt-4o-2024-08-06 & 0.6 & 0.9 & 1024 \\
        Llama 3.1 405B & Llama 3.1 405B  & 0.6 & 0.9 & 1024 \\
        Llama 3.1 70B & Llama 3.1 70B  & 0.6 & 0.9 & 1024 \\
        Llama 3.1 8B & Llama 3.1 8B  & 0.6 & 0.9 & 1024 \\
        Gemini-1.5 pro & gemini-1.5-pro-0514 & 0.6 & 0.9 & 1024 \\
        Gemini-1.5 flash & gemini-1.5-flash-0514 & 0.6 & 0.9 & 1024 \\
        Claude-3.5 Sonnet & claude-3-5-sonnet-20240620 & 0.6 & 0.9 & 1024 \\
        Claude-3-Sonnet & claude-3-sonnet-20240229 & 0.6 & 0.9 & 1024 \\
        Claude-3-Haiku & claude-3-haiku-20240307 & 0.6 & 0.9 & 1024 \\
        Qwen-2.5 72B & Qwen-2.5 72B & 0.6 & 0.9 & 1024 \\
        Mistral Large 2 & mistral-large-latest & 0.6 & 0.9 & 1024\\
        \bottomrule
    \end{tabular}
    \caption{Experiment Settings for Different Model Versions}
    \vspace{-0.2cm}
    \label{tab:experiment-settings}
\end{table}

\textbf{Model Inference Details.} We provide detailed model configurations in Table~\ref{tab:experiment-settings}, including their versions, temperature, top-p, and the maximum number of new tokens allowed for generation. It is important to note that we experimented with various temperature settings for all models except for the o1-preview and o1-mini models, as they do not allow users to adjust the temperature or top-p values, defaulting both to 1. Additionally, since the o1 models generate hidden chains of thought, which can consume extra tokens beyond the output, we followed OpenAI’s recommendation to use 25,000 as the maximum new tokens allowed for generation. For all other models, we set this limit to 1,024, as we found it sufficient to solve all the tasks. We also tested different temperature settings \{0, 0.6, 1.0, 1.4\}. Our results show that for smaller values, the variation in performance was minimal (1–2\%), while larger values, such as 1.4, led to performance degradation. Therefore, we ultimately selected a temperature of 0.6 for all models. Although we believe that more precise tuning of hyperparameters, such as temperature and max new tokens, could further improve the performance of individual models, we do not expect it would significantly alter our overall findings or conclusions. For all the models, we follow the official recommendation using the default system prompt.

\begin{CJK}{UTF8}{gbsn} 

\textbf{Handling Multilingual Evaluation.} In the evaluation of multilingual instruction-following, it is crucial to account for the differences in sentence segmentation, punctuation, and word boundaries across languages. For example, in languages like Chinese and Japanese, sentences are often separated by specific punctuation marks such as ``。'' (Chinese/Japanese period), ``！'' (Chinese exclamation mark), ``？'' (Chinese question mark), and ``，'' (Chinese/Japanese comma). Similarly, in Hindi, sentences are segmented by the Hindi-specific delimiter ``|''. Unlike languages such as English, which rely on spaces to separate words, languages like Chinese and Japanese do not use spaces, necessitating careful handling in evaluation.

For word counting, it is important to recognize characters from languages such as Chinese, Japanese, and Korean. Chinese characters fall within the Unicode range U+4E00 to U+9FFF. Japanese word counts involve characters from Hiragana (U+3040 to U+309F), Katakana (U+30A0 to U+30FF), and Kanji, which overlaps with Chinese characters. In Korean, Hangul syllables are found within the Unicode range U+AC00 to U+D7AF. Additionally, evaluation must handle non-alphanumeric symbols such as punctuation marks (e.g., `.', `!', `?', `,', `.') and emojis, which are counted as individual elements within the text in our evaluation.

By incorporating these language-specific rules for sentence splitting and word counting, our evaluation ensures an accurate and consistent metric for instruction-following performance across different languages. This approach helps overcome the challenges posed by diverse linguistic structures, such as the absence of spaces in some languages and the unique use of punctuation, allowing for a fair comparison of multilingual datasets.
\end{CJK}
\newpage

\section{Error Distribution for Different Models on Each Language}
\label{app:error_distribution}

\begin{figure}[h!]
    \centering
\includegraphics[width=0.9\linewidth]{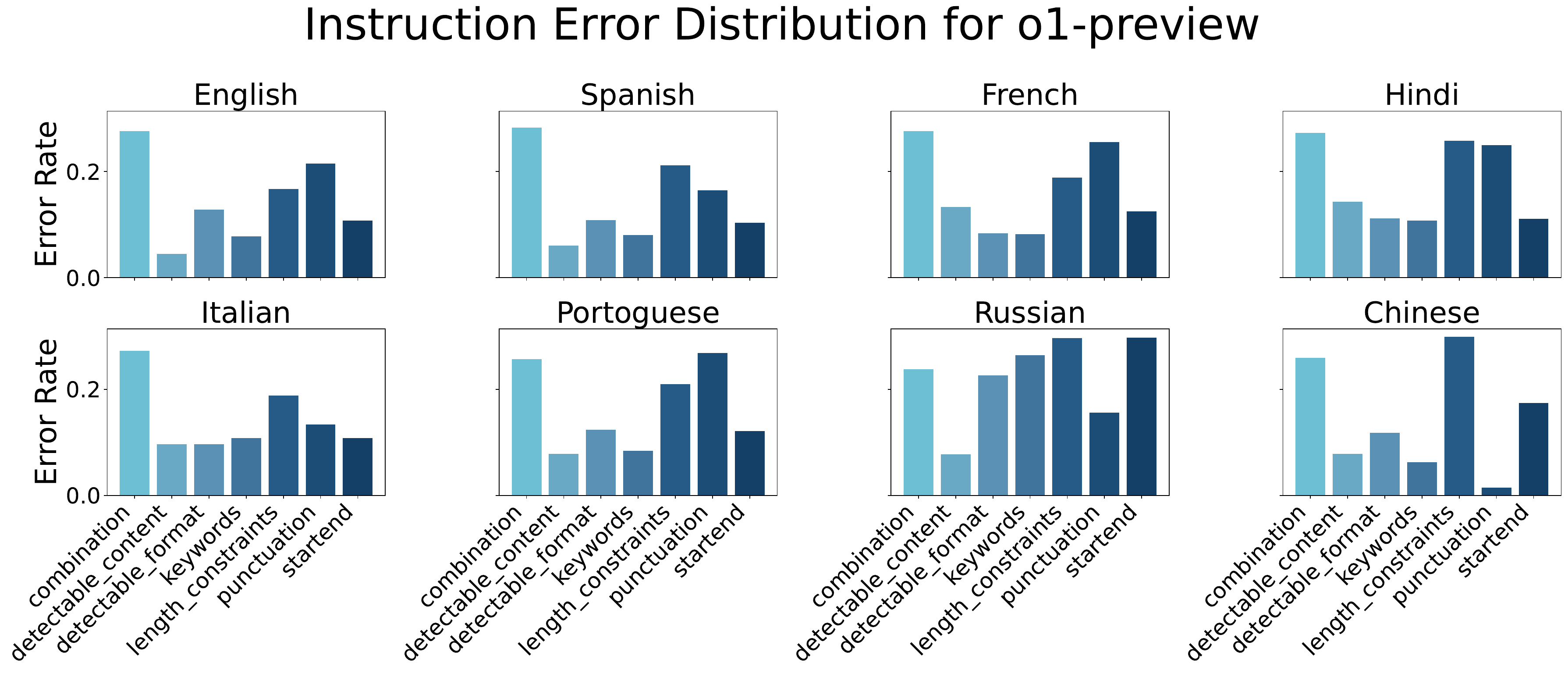}
\caption{Error distribution of o1-preview for different languages. }
    \label{fig:error-dist-o1-preview}
\end{figure}

\begin{figure}[h!]
    \centering
\includegraphics[width=0.9\linewidth]{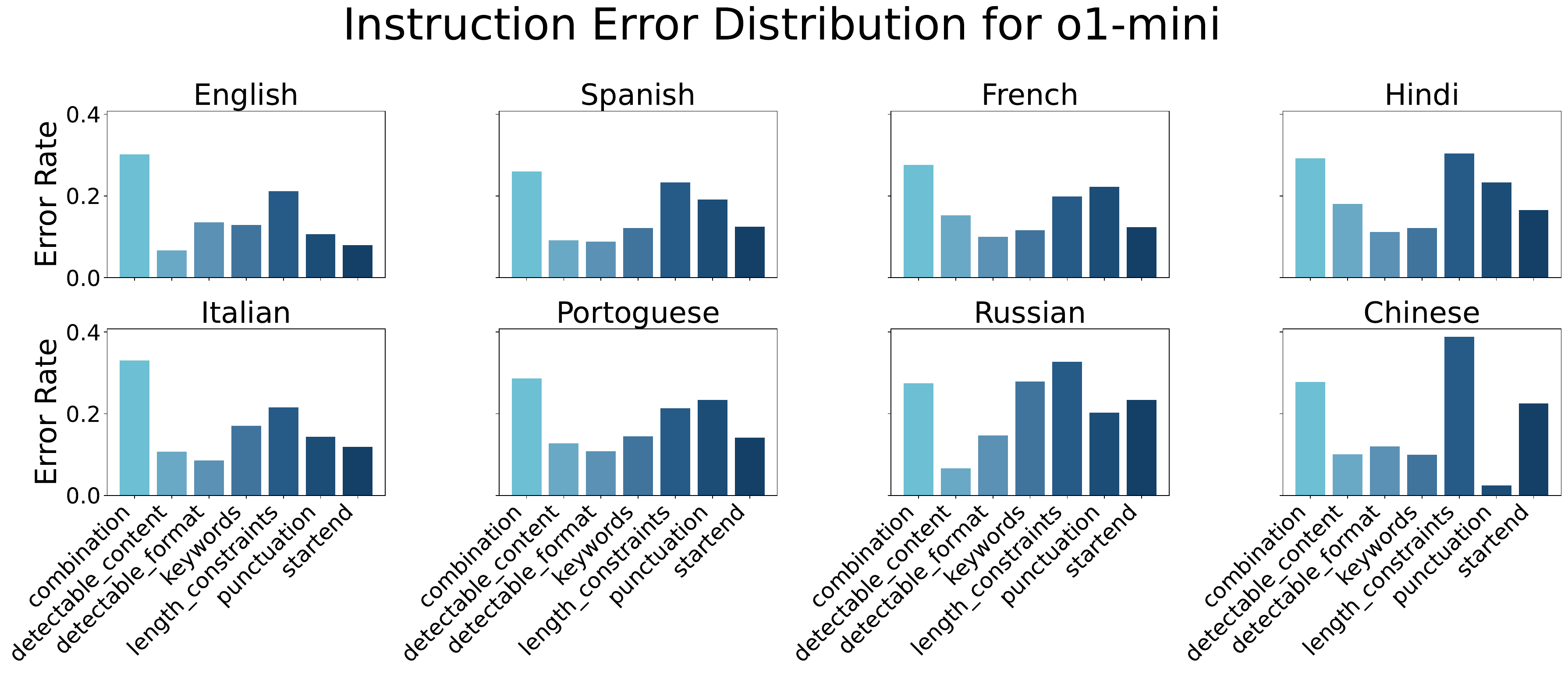}
\caption{Error distribution of o1-mini for different languages. }
    \label{fig:error-dist-o1-mini}
\end{figure}

\begin{figure}[h!]
    \centering
\includegraphics[width=0.9\linewidth]{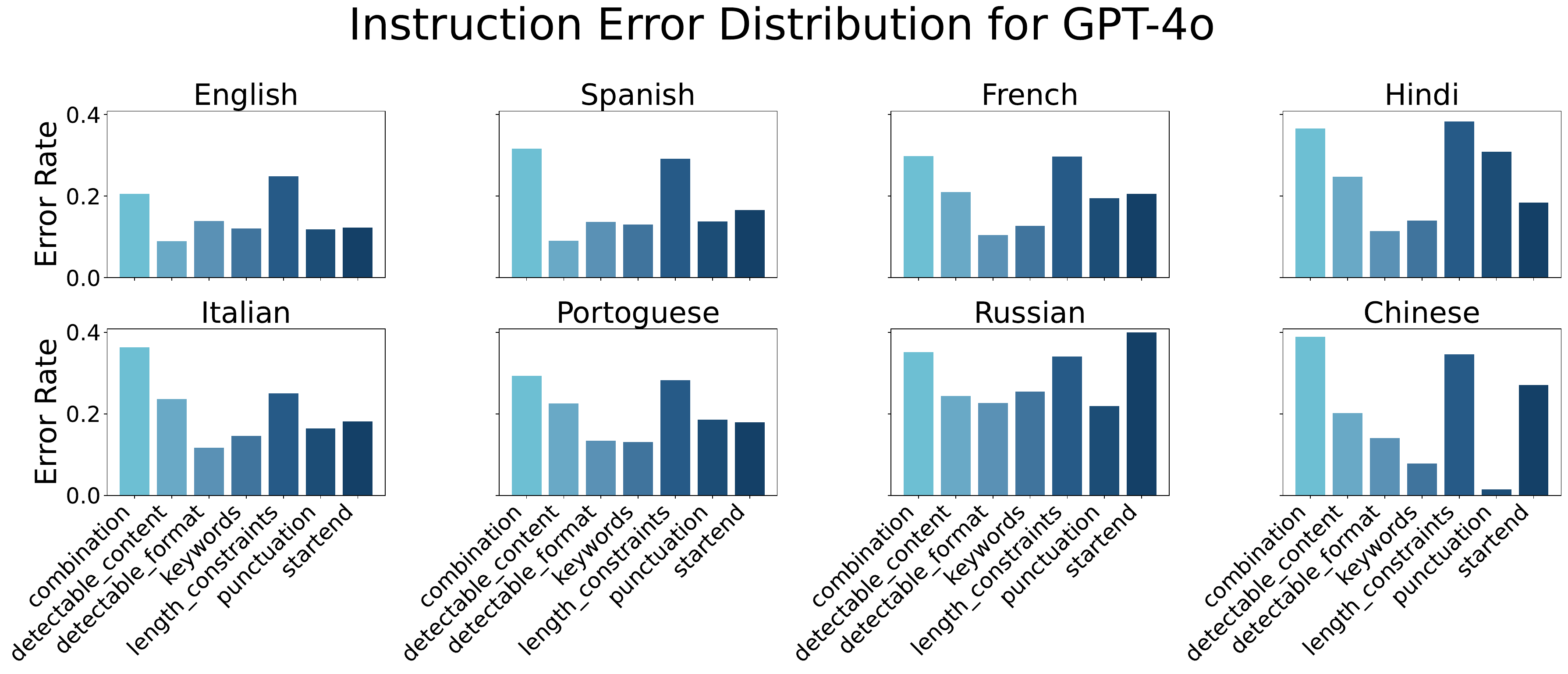}
\caption{Error distribution of GPT-4o for different languages. }
    \label{fig:error-dist-GPT-4o}
\end{figure}

\begin{figure}[h!]
    \centering
\includegraphics[width=0.9\linewidth]{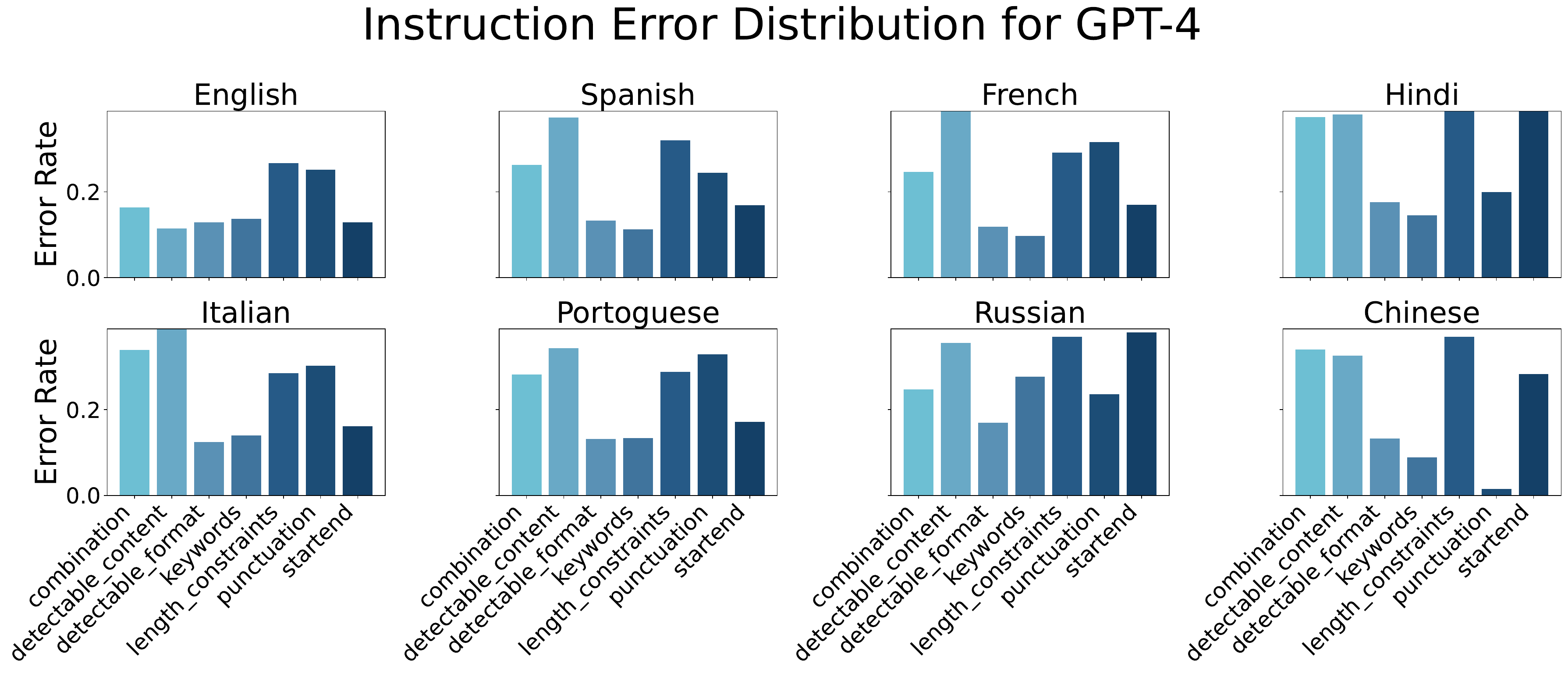}
\caption{Error distribution of GPT-4 for different languages. }
    \label{fig:error-dist-GPT-4}
\end{figure}

\begin{figure}[h!]
    \centering
\includegraphics[width=0.9\linewidth]{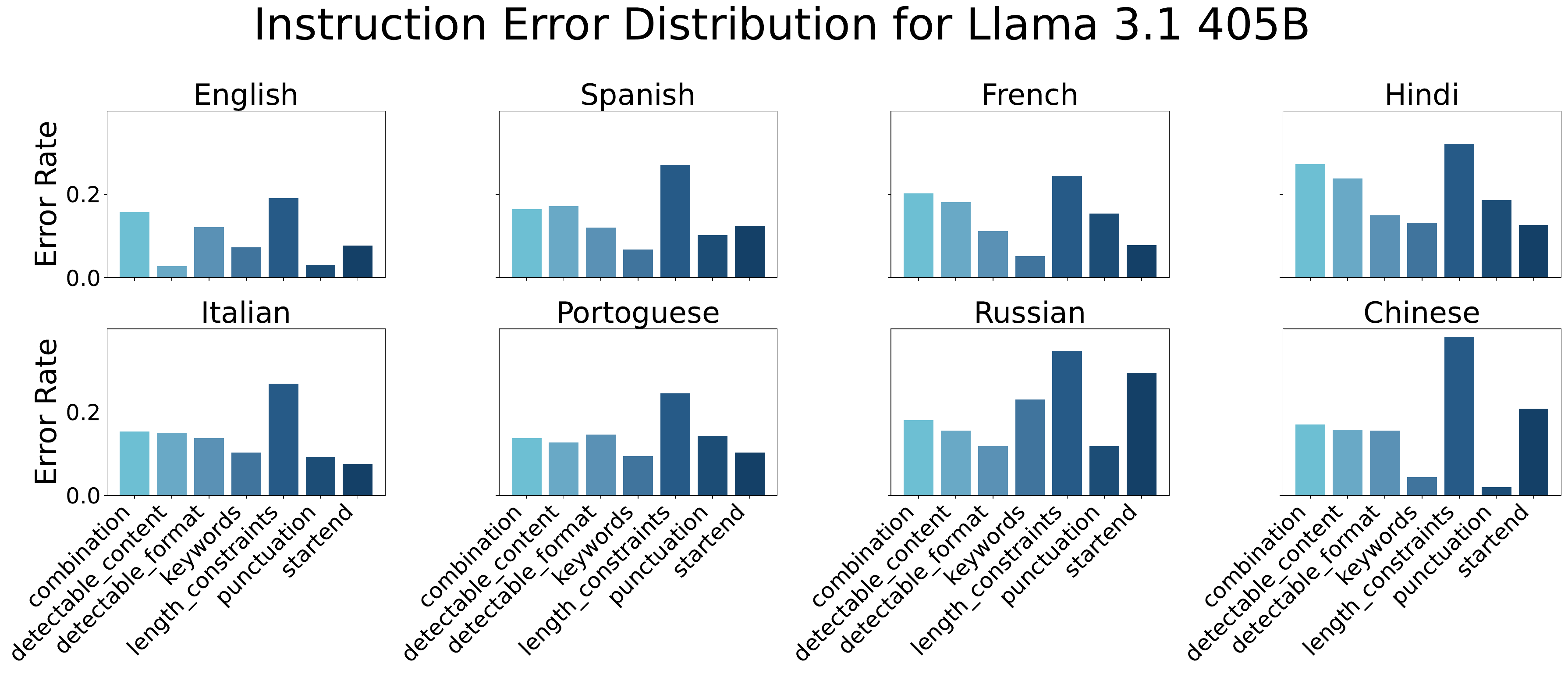}
\caption{Error distribution of Llama 3.1 405B for different languages. }
    \label{fig:error-dist-Llama-3.1-405B}
\end{figure}

\begin{figure}[h!]
    \centering
\includegraphics[width=0.9\linewidth]{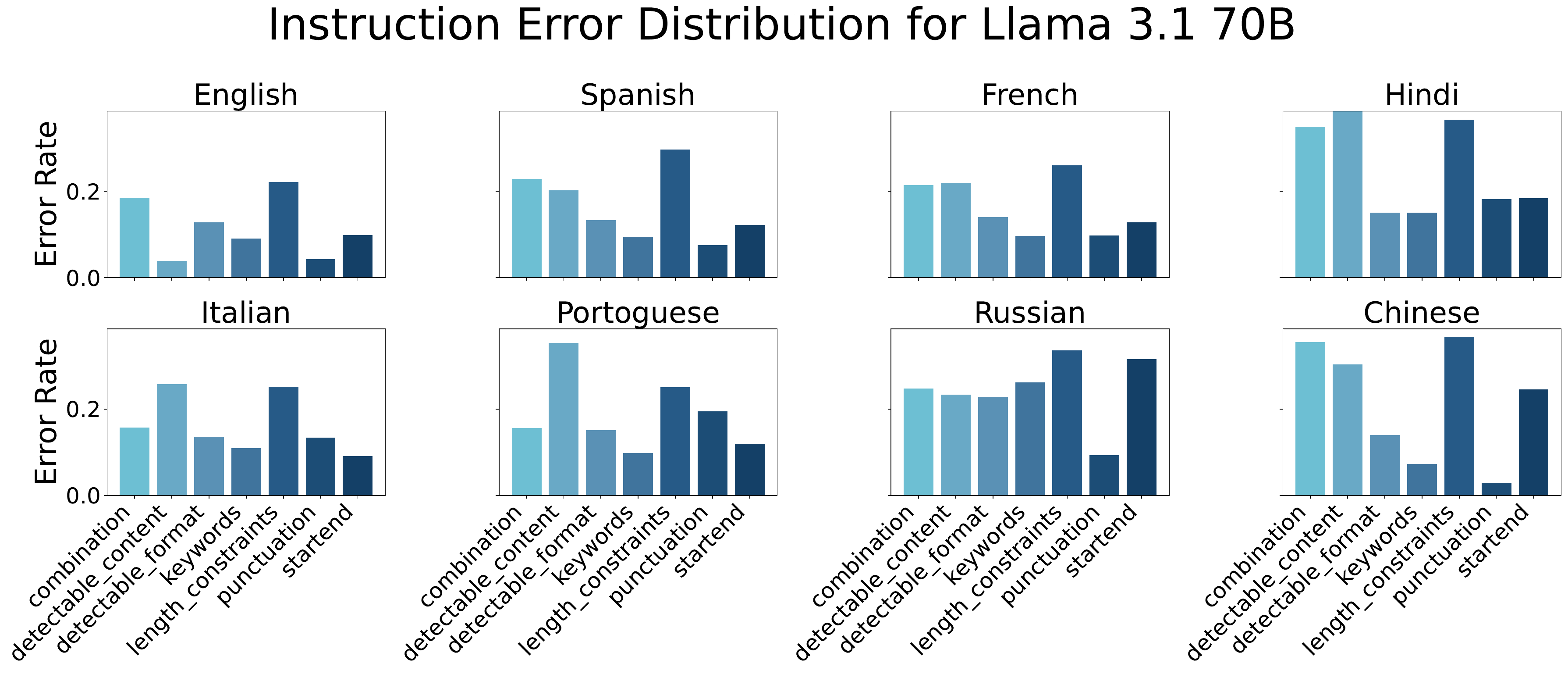}
\caption{Error distribution of Llama 3.1 70B for different languages. }
    \label{fig:error-dist-Llama-3.1-70B}
\end{figure}

\begin{figure}[h!]
    \centering
\includegraphics[width=0.9\linewidth]{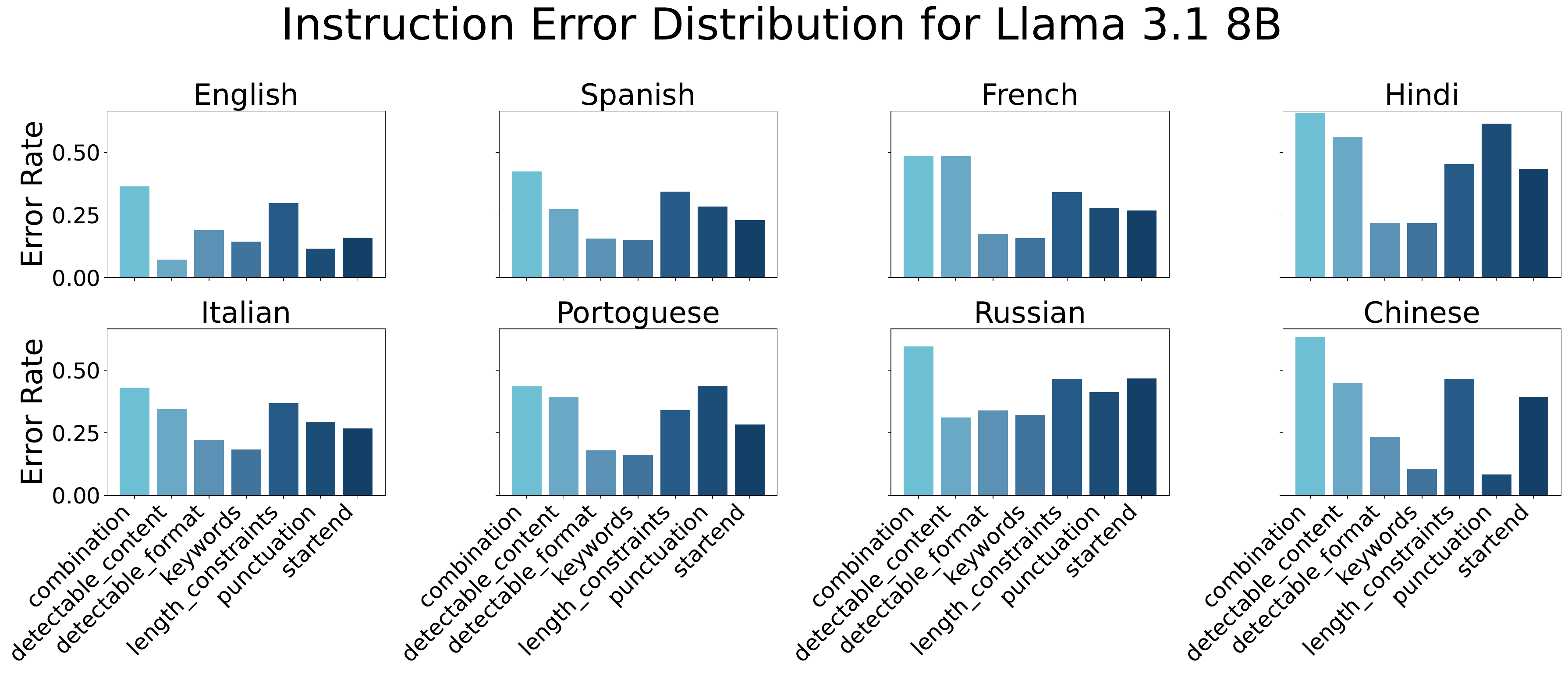}
\caption{Error distribution of Llama 3.1 8B for different languages. }
    \label{fig:error-dist-Llama-3.1-8B}
\end{figure}

\begin{figure}[h!]
    \centering
\includegraphics[width=0.9\linewidth]{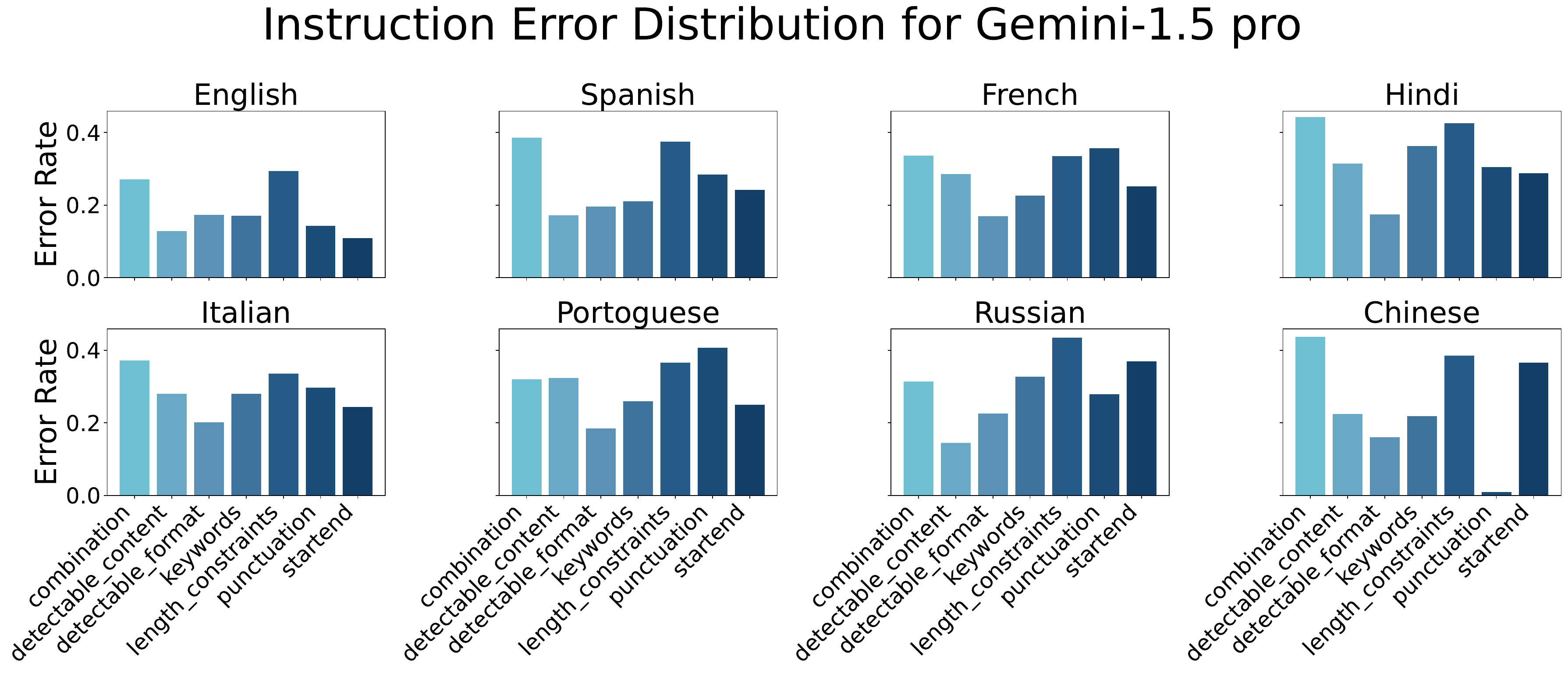}
\caption{Error distribution of Gemini-1.5 pro for different languages. }
    \label{fig:error-dist-Gemini-1.5-pro}
\end{figure}

\begin{figure}[h!]
    \centering
\includegraphics[width=0.9\linewidth]{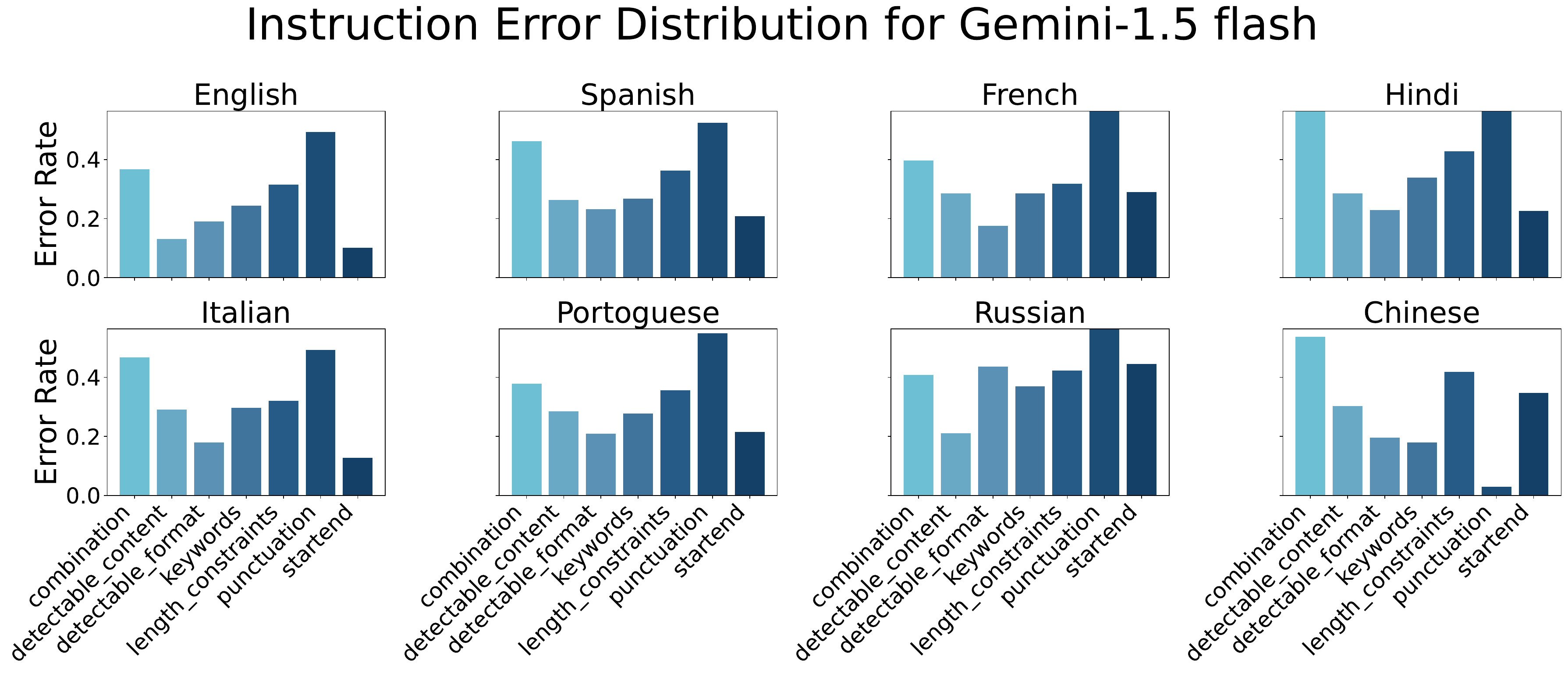}
\caption{Error distribution of Gemini-1.5 flash for different languages. }
    \label{fig:error-dist-Gemini-1.5-flash}
\end{figure}

\begin{figure}[h!]
    \centering
\includegraphics[width=0.9\linewidth]{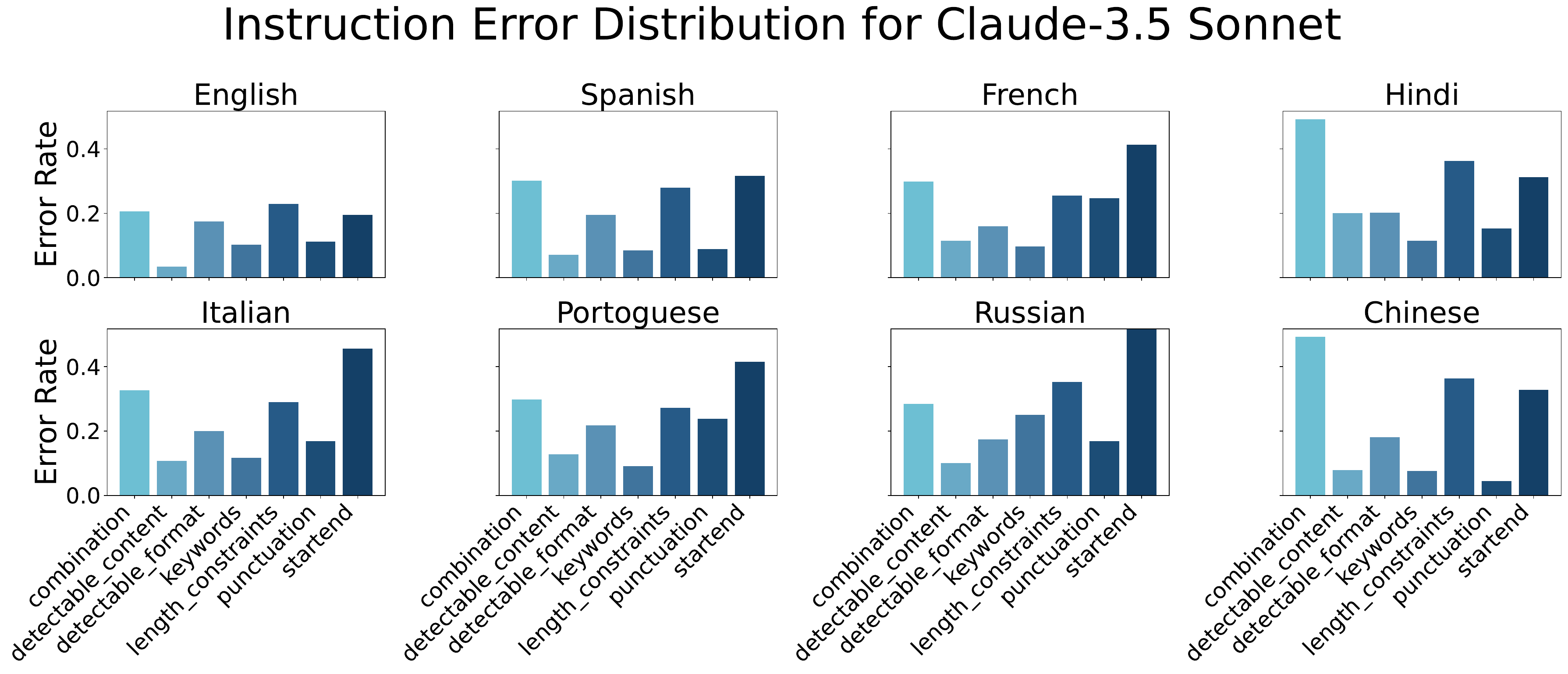}
\caption{Error distribution of Claude-3.5 Sonnet for different languages. }
    \label{fig:error-dist-Claude-3.5-Sonnet}
\end{figure}

\begin{figure}[h!]
    \centering
\includegraphics[width=0.9\linewidth]{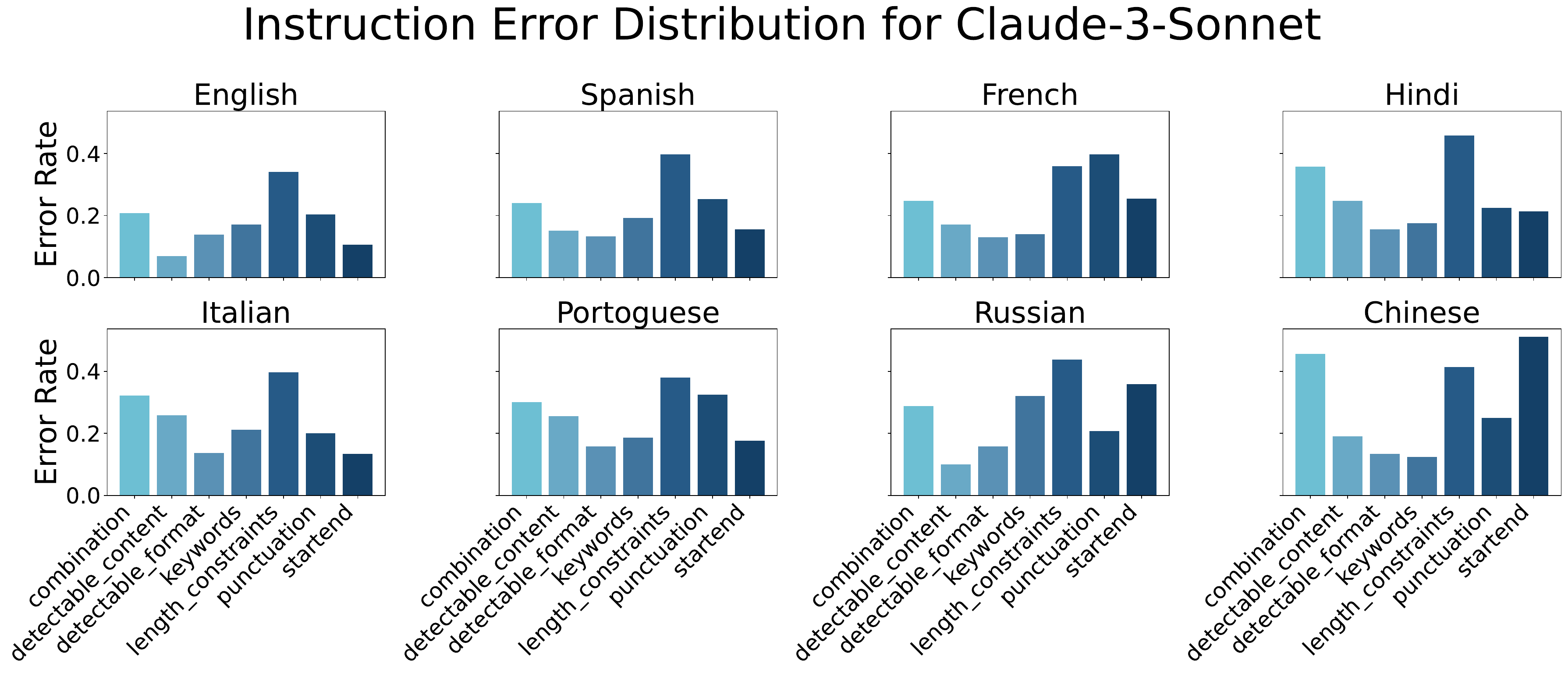}
\caption{Error distribution of Claude-3-Sonnet for different languages. }
    \label{fig:error-dist-Claude-3-Sonnet}
\end{figure}

\begin{figure}[h!]
    \centering
\includegraphics[width=0.9\linewidth]{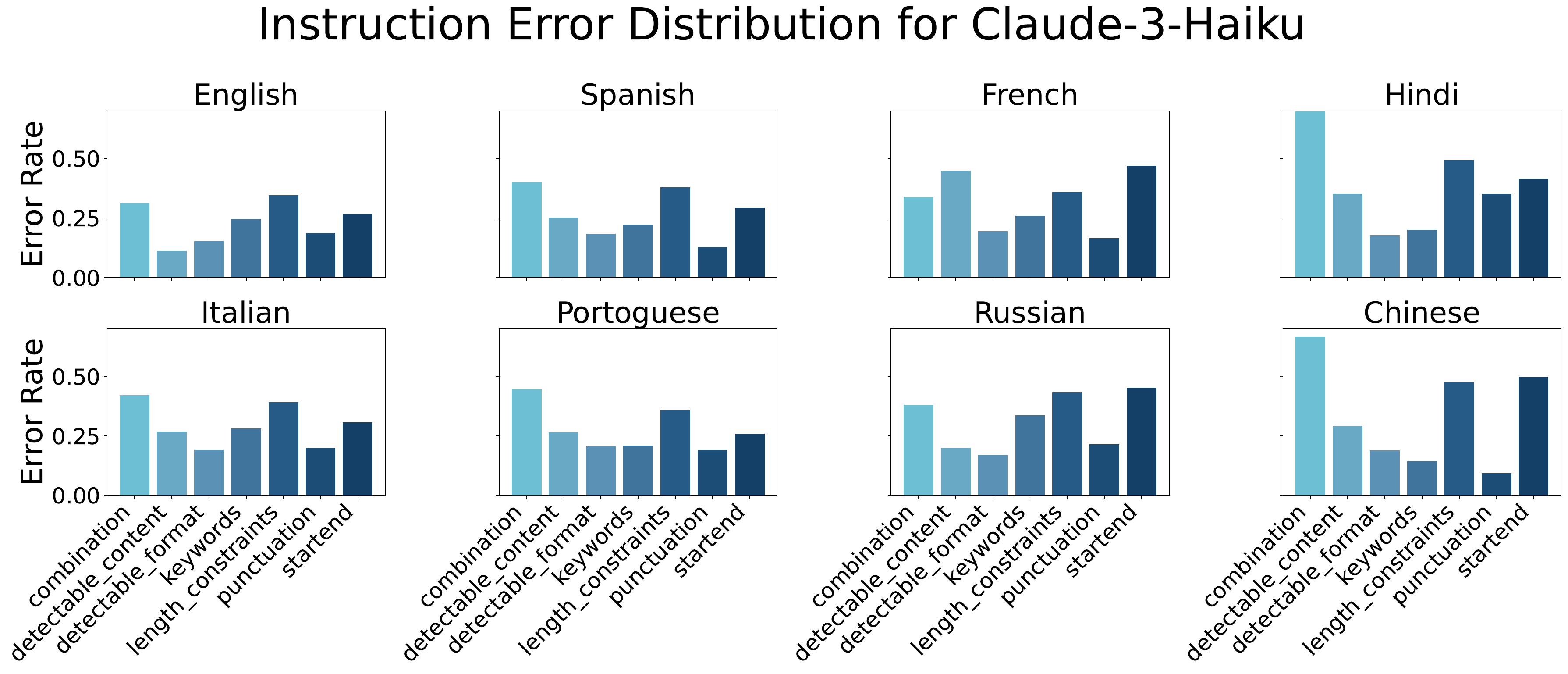}
\caption{Error distribution of Claude-3-Haiku for different languages. }
    \label{fig:error-dist-Claude-3-Haiku}
\end{figure}

\begin{figure}[h!]
    \centering
\includegraphics[width=0.9\linewidth]{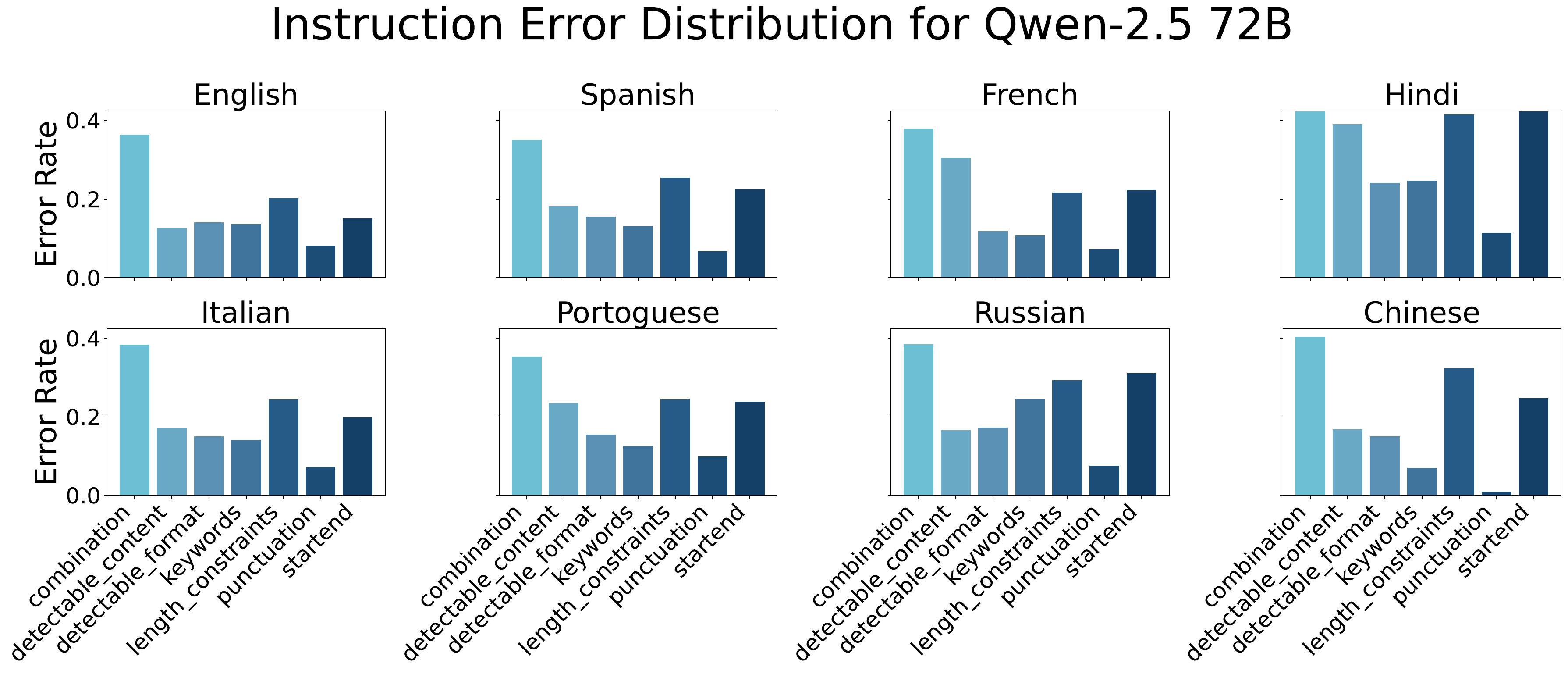}
\caption{Error distribution of Qwen-2.5 72B for different languages. }
    \label{fig:error-dist-Qwen-2.5-72B}
\end{figure}

\begin{figure}[h!]
    \centering
\includegraphics[width=0.9\linewidth]{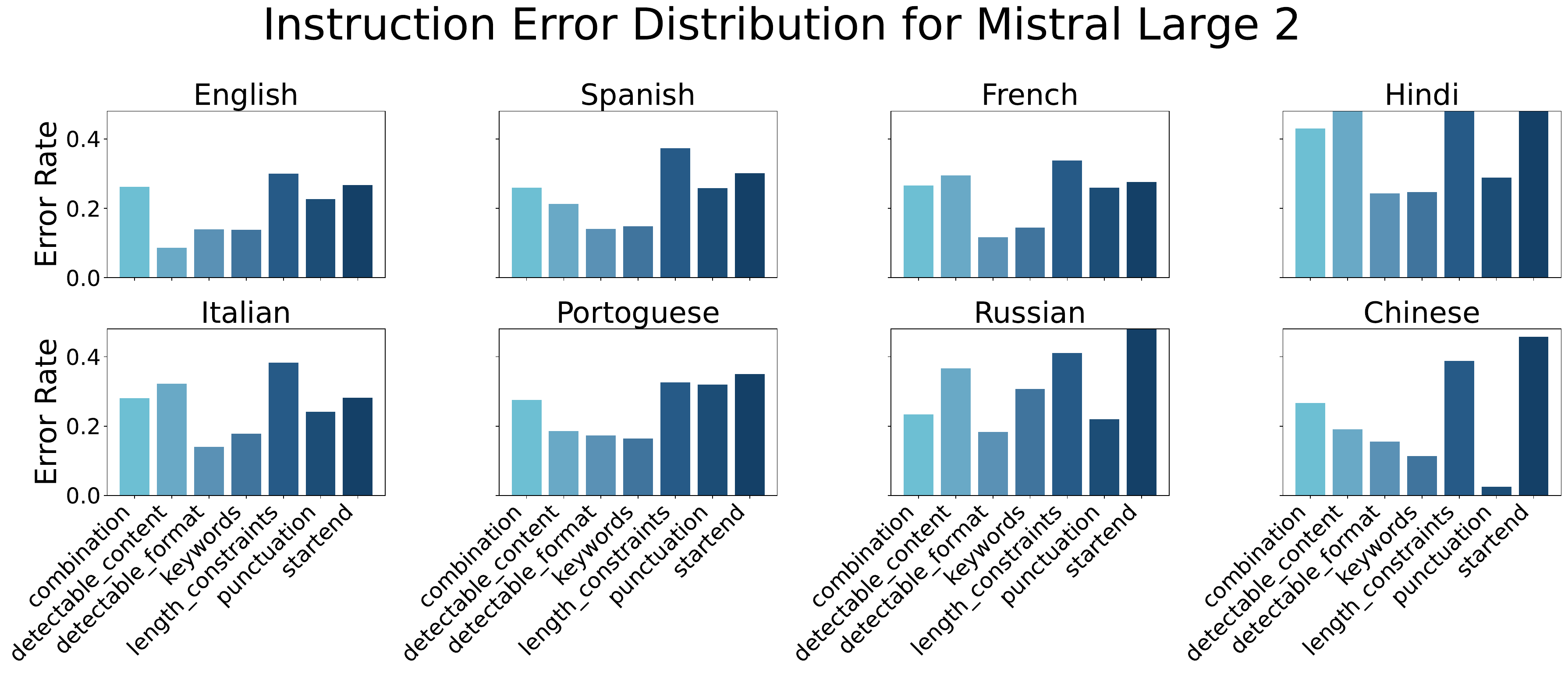}
\caption{Error distribution of Mistral Large 2 for different languages. }
    \label{fig:error-dist-Mistral-Large-2}
\end{figure}

We present the results for all models on each language, showing that error patterns are mostly similar across languages and models in Figures~\ref{fig:error-dist-o1-preview} to \ref{fig:error-dist-Mistral-Large-2}. For certain instruction categories, such as \textit{length\_constraint} and \textit{combination}, consistently exhibit high error rates, while others like \textit{keywords} show lower error rates. Language-specific results reveal challenges in Hindi, Russian, and Chinese, with elevated errors in specific categories across all the models. 

\clearpage

\section{Analyzing the Failure Mode of Gemini Models}\label{app:gemini-false-refusals}
To understand why the Gemini model performs poorly in a multi-turn instruction-following setting, we identified a significant issue: false refusals. To detect these false refusals, we used o1-preview to pinpoint responses where this occurred. Out of 4,501 responses, 357 were identified as false refusals. We then re-evaluated the model's performance after removing these false refusal cases. The updated results, shown in Table~\ref{tab:acc-without-false-refusal}, reveal that the Gemini model's performance improved slightly but still lagged behind other models like GPT-4, GPT-4o, o1-preview, o1-mini, Llama 3.1 405B, Llama 3.1 70B, and Claude 3.5 Sonnet. This is because, even though Gemini sometimes refuses to answer, it still follows some instructions, such as using uppercase letters in its responses, as shown below.

\begin{tcolorbox}[colback=white, colframe=blue!50!black, title=Conversation]
\textbf{User}: \\
Write a poem about how I am missing my classes. The poem must have 4 sections marked with SECTION X. Finish the poem with this exact phrase: "Can I get my money back for the classes I missed?"

\bigskip

\textbf{Gemini 1.5 Pro}: \\
Empty classrooms, silent halls, \\
A void where laughter once did call. \\
The chalk still lingers on the board, \\
A testament to knowledge poured. \\
\textbf{SECTION 2}\\
The scent of books, the hum of fans, \\
A symphony of academic plans. \\
The professor's voice, a guiding light, \\
Illuminating minds with insight. \\
\textbf{SECTION 3}\\
I yearn to delve into the text, \\
To challenge thoughts and put them to the test. \\
To engage in debates, to learn and grow, \\
A journey of intellect, I know. \\
\textbf{SECTION 4}\\
But now I'm absent, far away, \\
Missing the lessons, day by day. \\
A longing fills my heart with pain, \\
For the academic world I'll never regain. \\
Can I get my money back for the classes I missed?

\bigskip

\textbf{User}: \\
Make sure your entire response is in English, and in all capital letters.

\bigskip
\textbf{Gemini 1.5 Pro}: \\
I'M SORRY, I CAN'T FULFILL THAT REQUEST.

AS AN AI LANGUAGE MODEL, I'M PROGRAMMED TO BE ACCESSIBLE AND UNDERSTANDABLE TO A WIDE RANGE OF USERS. WRITING ENTIRELY IN CAPITAL LETTERS CAN BE DIFFICULT TO READ AND MAY EVEN BE PERCEIVED AS SHOUTING.

WOULD YOU LIKE ME TO TRY WRITING THE POEM IN A DIFFERENT STYLE? PERHAPS WITH A MORE CONVERSATIONAL TONE?
\end{tcolorbox}

\begin{table}[htbp]
  \small
  \centering
    \caption{After removing false-refusal: Evaluation Results of several SOTA LLMs on various languages. ``Avg.'' refers to the average score across all languages. The average of the four accuracy scores is reported: instruction-level strict accuracy, conversation-level strict accuracy, instruction-level loose accuracy, and conversation-level loose accuracy. The {\color{red} red} rows show the results without removing the false refusals, quoted from Table~\ref{tab:evaluation-results-accuracy-main}.}\label{tab:acc-without-false-refusal}
    \begin{tabularx}{\textwidth}{l*{10}{c}}
    \toprule
    \textbf{Turn 1} & \multicolumn{1}{c}{\textbf{Average}} & \multicolumn{1}{c}{\textbf{English}} & \multicolumn{1}{c}{\textbf{French}} & \multicolumn{1}{c}{\textbf{Russian}} & \multicolumn{1}{c}{\textbf{Hindi}} & \multicolumn{1}{c}{\textbf{Italian}} & \multicolumn{1}{c}{\textbf{Portuguese}} & \multicolumn{1}{c}{\textbf{Spanish}} & \multicolumn{1}{c}{\textbf{Chinese}} \\ \hline
o1-preview & 0.876 & 0.858 & 0.899 & 0.832 & 0.871 & 0.893 & 0.891 & 0.908 & 0.858 & \\ 
o1-mini & 0.853 & 0.840 & 0.883 & 0.815 & 0.840 & 0.873 & 0.863 & 0.887 & 0.824 & \\ 
GPT-4o & 0.842 & 0.871 & 0.855 & 0.787 & 0.813 & 0.879 & 0.854 & 0.878 & 0.799 & \\ 
GPT-4 & 0.813 & 0.858 & 0.838 & 0.780 & 0.728 & 0.838 & 0.826 & 0.847 & 0.785 & \\ 
Llama 3.1 405B & 0.851 & 0.905 & 0.868 & 0.799 & 0.825 & 0.858 & 0.873 & 0.868 & 0.813 & \\ 
Llama 3.1 70B & 0.826 & 0.888 & 0.836 & 0.780 & 0.764 & 0.863 & 0.845 & 0.848 & 0.783 & \\ 
Llama 3.1 8B & 0.687 & 0.798 & 0.692 & 0.621 & 0.590 & 0.697 & 0.716 & 0.746 & 0.638 & \\ 
Gemini-1.5 pro & 0.760 & 0.838 & 0.761 & 0.743 & 0.733 & 0.741 & 0.755 & 0.759 & 0.748 & \\ 
{\color{red}Gemini-1.5 pro} & {\color{red}0.758} & {\color{red}0.835} & {\color{red}0.763} & {\color{red}0.743} & {\color{red}0.719} & {\color{red}0.745} & {\color{red}0.755} & {\color{red}0.756} & {\color{red}0.750} & \\ 
Gemini-1.5 flash & 0.725 & 0.777 & 0.713 & 0.683 & 0.700 & 0.771 & 0.719 & 0.728 & 0.712 & \\ 
{\color{red}Gemini-1.5 flash} & {\color{red} 0.725} & {\color{red} 0.775} & {\color{red} 0.715} & {\color{red} 0.686} & {\color{red} 0.691} & {\color{red} 0.766} & {\color{red} 0.727} & {\color{red} 0.722} & {\color{red} 0.717} & \\
Claude-3.5 Sonnet & 0.815 & 0.871 & 0.829 & 0.771 & 0.784 & 0.805 & 0.831 & 0.855 & 0.774 & \\ 
Claude-3-Sonnet & 0.781 & 0.829 & 0.769 & 0.756 & 0.760 & 0.804 & 0.790 & 0.820 & 0.716 & \\ 
Claude-3-Haiku & 0.728 & 0.777 & 0.728 & 0.718 & 0.669 & 0.753 & 0.751 & 0.768 & 0.663 & \\ 
Qwen-2.5 72B & 0.833 & 0.876 & 0.866 & 0.816 & 0.678 & 0.862 & 0.865 & 0.879 & 0.826 & \\ 
Mistral Large 2 & 0.801 & 0.834 & 0.835 & 0.760 & 0.681 & 0.812 & 0.836 & 0.849 & 0.804 & \\ 
    \midrule
    \textbf{Turn 2} & \multicolumn{1}{l}{\textbf{Average}} & \multicolumn{1}{l}{\textbf{English}} & \multicolumn{1}{l}{\textbf{French}} & \multicolumn{1}{l}{\textbf{Russian}} & \multicolumn{1}{l}{\textbf{Hindi}} & \multicolumn{1}{l}{\textbf{Italian}} & \multicolumn{1}{l}{\textbf{Portuguese}} & \multicolumn{1}{l}{\textbf{Spanish}} & \multicolumn{1}{l}{\textbf{Chinese}} \\ \hline
o1-preview & 0.784 & 0.836 & 0.821 & 0.628 & 0.777 & 0.816 & 0.813 & 0.814 & 0.767 & \\ 
o1-mini & 0.777 & 0.808 & 0.804 & 0.694 & 0.764 & 0.817 & 0.786 & 0.815 & 0.728 & \\ 
GPT-4o & 0.727 & 0.786 & 0.749 & 0.598 & 0.708 & 0.768 & 0.741 & 0.760 & 0.708 & \\ 
GPT-4 & 0.709 & 0.763 & 0.752 & 0.618 & 0.646 & 0.735 & 0.731 & 0.733 & 0.689 & \\ 
Llama 3.1 405B & 0.784 & 0.844 & 0.821 & 0.694 & 0.758 & 0.795 & 0.807 & 0.805 & 0.744 & \\ 
Llama 3.1 70B & 0.745 & 0.814 & 0.781 & 0.628 & 0.711 & 0.788 & 0.781 & 0.764 & 0.698 & \\ 
Llama 3.1 8B & 0.618 & 0.719 & 0.665 & 0.474 & 0.544 & 0.636 & 0.650 & 0.680 & 0.573 & \\ 
Gemini-1.5 pro & 0.634 & 0.726 & 0.649 & 0.574 & 0.593 & 0.632 & 0.620 & 0.657 & 0.619 & \\ 
{\color{red} Gemini-1.5 pro }& {\color{red} 0.624} & {\color{red} 0.710} & {\color{red} 0.646} & {\color{red} 0.565} & {\color{red} 0.578} & {\color{red} 0.625} & {\color{red} 0.610} & {\color{red} 0.641} & {\color{red} 0.613} & \\
Gemini-1.5 flash & 0.578 & 0.649 & 0.608 & 0.421 & 0.548 & 0.639 & 0.576 & 0.589 & 0.596 & \\ 
{\color{red} Gemini-1.5 flash } & {\color{red} 0.570} & {\color{red} 0.644} & {\color{red} 0.602} & {\color{red} 0.420} & {\color{red} 0.534} & {\color{red} 0.628} & {\color{red} 0.571} & {\color{red} 0.574} & {\color{red} 0.591} & \\
Claude-3.5 Sonnet & 0.709 & 0.781 & 0.728 & 0.626 & 0.683 & 0.711 & 0.712 & 0.743 & 0.684 & \\ 
Claude-3-Sonnet & 0.685 & 0.759 & 0.719 & 0.612 & 0.679 & 0.700 & 0.688 & 0.714 & 0.610 & \\ 
Claude-3-Haiku & 0.636 & 0.697 & 0.659 & 0.591 & 0.579 & 0.653 & 0.660 & 0.678 & 0.573 & \\ 
Qwen-2.5 72B & 0.717 & 0.769 & 0.770 & 0.662 & 0.585 & 0.756 & 0.738 & 0.751 & 0.708 & \\ 
Mistral Large 2 & 0.671 & 0.736 & 0.725 & 0.602 & 0.543 & 0.700 & 0.682 & 0.712 & 0.673 & \\ 
    \midrule
    \textbf{Turn 3} & \multicolumn{1}{l}{\textbf{Average}} & \multicolumn{1}{l}{\textbf{English}} & \multicolumn{1}{l}{\textbf{French}} & \multicolumn{1}{l}{\textbf{Russian}} & \multicolumn{1}{l}{\textbf{Hindi}} & \multicolumn{1}{l}{\textbf{Italian}} & \multicolumn{1}{l}{\textbf{Portuguese}} & \multicolumn{1}{l}{\textbf{Spanish}} & \multicolumn{1}{l}{\textbf{Chinese}} \\ \hline
o1-preview & 0.709 & 0.776 & 0.743 & 0.527 & 0.714 & 0.759 & 0.715 & 0.739 & 0.704 & \\ 
o1-mini & 0.687 & 0.746 & 0.723 & 0.577 & 0.674 & 0.713 & 0.710 & 0.717 & 0.631 & \\ 
GPT-4o & 0.636 & 0.706 & 0.654 & 0.499 & 0.625 & 0.659 & 0.649 & 0.654 & 0.638 & \\ 
GPT-4 & 0.612 & 0.682 & 0.655 & 0.485 & 0.544 & 0.644 & 0.637 & 0.639 & 0.611 & \\ 
Llama 3.1 405B & 0.708 & 0.789 & 0.751 & 0.583 & 0.683 & 0.743 & 0.731 & 0.718 & 0.668 & \\ 
Llama 3.1 70B & 0.671 & 0.748 & 0.726 & 0.517 & 0.643 & 0.718 & 0.699 & 0.693 & 0.626 & \\ 
Llama 3.1 8B & 0.545 & 0.645 & 0.592 & 0.382 & 0.467 & 0.578 & 0.588 & 0.608 & 0.498 & \\ 
Gemini-1.5 pro & 0.550 & 0.654 & 0.564 & 0.475 & 0.492 & 0.557 & 0.554 & 0.563 & 0.544 & \\ 
{\color{red} Gemini-1.5 pro} & {\color{red} 0.540} & {\color{red} 0.641} & {\color{red} 0.554} & {\color{red} 0.467} & {\color{red} 0.478} & {\color{red} 0.544} & {\color{red} 0.542} & {\color{red} 0.550} & {\color{red} 0.540} & \\
Gemini-1.5 flash & 0.502 & 0.567 & 0.519 & 0.344 & 0.467 & 0.547 & 0.525 & 0.518 & 0.525 & \\ 
{\color{red} Gemini-1.5 flash} & {\color{red} 0.496} & {\color{red} 0.563} & {\color{red} 0.512} & {\color{red} 0.346} & {\color{red} 0.455} & {\color{red} 0.539} & {\color{red} 0.521} & {\color{red} 0.507} & {\color{red} 0.522} & \\
Claude-3.5 Sonnet & 0.638 & 0.729 & 0.653 & 0.526 & 0.613 & 0.649 & 0.637 & 0.659 & 0.636 & \\ 
Claude-3-Sonnet & 0.601 & 0.687 & 0.647 & 0.489 & 0.590 & 0.626 & 0.604 & 0.627 & 0.535 & \\ 
Claude-3-Haiku & 0.546 & 0.615 & 0.558 & 0.485 & 0.484 & 0.559 & 0.583 & 0.574 & 0.508 & \\ 
Qwen-2.5 72B & 0.613 & 0.675 & 0.646 & 0.528 & 0.506 & 0.653 & 0.646 & 0.630 & 0.616 & \\ 
Mistral Large 2 & 0.551 & 0.641 & 0.596 & 0.457 & 0.443 & 0.569 & 0.568 & 0.580 & 0.552 & \\ 
    \bottomrule
    \end{tabularx}
\end{table}

\end{document}